\documentclass[10pt,twocolumn,letterpaper]{article}
\pdfoutput=1

\usepackage{iccv}
\usepackage{times}
\usepackage{epsfig}
\usepackage{graphicx}
\usepackage{amsmath}
\usepackage{amssymb}

\usepackage{booktabs}
\usepackage{multirow}
\usepackage{subfigure}
\usepackage[accsupp]{axessibility}  

\newcommand*{\affaddr}[1]{#1} 
\newcommand*{\affmark}[1][*]{\textsuperscript{#1}}
\renewcommand\thefootnote{}

\usepackage{enumitem}

\usepackage[pagebackref=true,breaklinks=true,letterpaper=true,colorlinks,bookmarks=false]{hyperref}

\iccvfinalcopy 

\def\iccvPaperID{3837} 

\ificcvfinal\fi

\begin{document}

\title{ALIP: Adaptive Language-Image Pre-training with Synthetic Caption}

\author{
Kaicheng Yang\affmark[1], 
Jiankang Deng\textsuperscript{*} \affmark[2],
Xiang An\affmark[1],  
Jiawei Li\affmark[1],
Ziyong Feng\affmark[1], \\ 
Jia Guo\affmark[3],   
Jing Yang\affmark[3],  
Tongliang Liu\affmark[4]\\
\affaddr{\affmark[1]DeepGlint} \qquad
\affaddr{\affmark[2]Huawei UKRD} \qquad
\affaddr{\affmark[3]InsightFace} \qquad 
\affaddr{\affmark[4]University of Sydney}\\
{\tt\small \{kaichengyang,xiangan,jiaweili,ziyongfeng\}@deepglint.com, tongliang.liu@sydney.edu.au} \\
{\tt\small \{jiankangdeng,guojia,y.jing2016\}@gmail.com}
}

\maketitle
\ificcvfinal\thispagestyle{empty}\fi

\begin{abstract}
 Contrastive Language-Image Pre-training (CLIP) has significantly boosted the performance of various vision-language tasks by scaling up the dataset with image-text pairs collected from the web. However, the presence of intrinsic noise and unmatched image-text pairs in web data can potentially affect the performance of representation learning. To address this issue, we first utilize the OFA model to generate synthetic captions that focus on the image content. The generated captions contain complementary information that is beneficial for pre-training. Then, we propose an Adaptive Language-Image Pre-training (ALIP), a bi-path model that integrates supervision from both raw text and synthetic caption. As the core components of ALIP, the Language Consistency Gate (LCG) and Description Consistency Gate (DCG) dynamically adjust the weights of samples and image-text/caption pairs during the training process. Meanwhile, the adaptive contrastive loss can effectively reduce the impact of noise data and enhances the efficiency of pre-training data. We validate ALIP with experiments on different scales of models and pre-training datasets. Experiments results show that ALIP achieves state-of-the-art performance on multiple downstream tasks including zero-shot image-text retrieval and linear probe. To facilitate future research, the code and pre-trained models are released at ~\url{https://github.com/deepglint/ALIP}.
 \footnote{\textsuperscript{*} corresponding author.} 
\setcounter{footnote}{0}
\renewcommand\thefootnote{\arabic{footnote}}
 
\end{abstract}

\section{Introduction}

With the development of mobile networks and social platforms, there has been an explosion in the production of image-text pairs on a massive scale~\cite{baltruvsaitis2018multimodal,guo2019deep}. This unprecedented abundance of data has laid a solid foundation for vision-language pre-training~\cite{radford2021learning,jia2021scaling}. Through image-text alignment on large-scale data, the Contrastive Language–Image Pre-training (CLIP) method~\cite{radford2021learning} has demonstrated huge success in multi-modal learning. Specifically, CLIP learns two separate unimodal encoders for image and text using a contrastive loss, one of the most effective losses for representation learning~\cite{tian2020contrastive,he2020momentum,chen2020simple,chopra2005learning}. Nevertheless, the negative impact of the noise in the web-crawled data has been largely overlooked, shadowed by the performance gain achieved from scaling up the training data~\cite{radenovic2023filtering,an2022unicom}.

\begin{figure}[t]
\centering
\includegraphics[width=1.0\columnwidth]{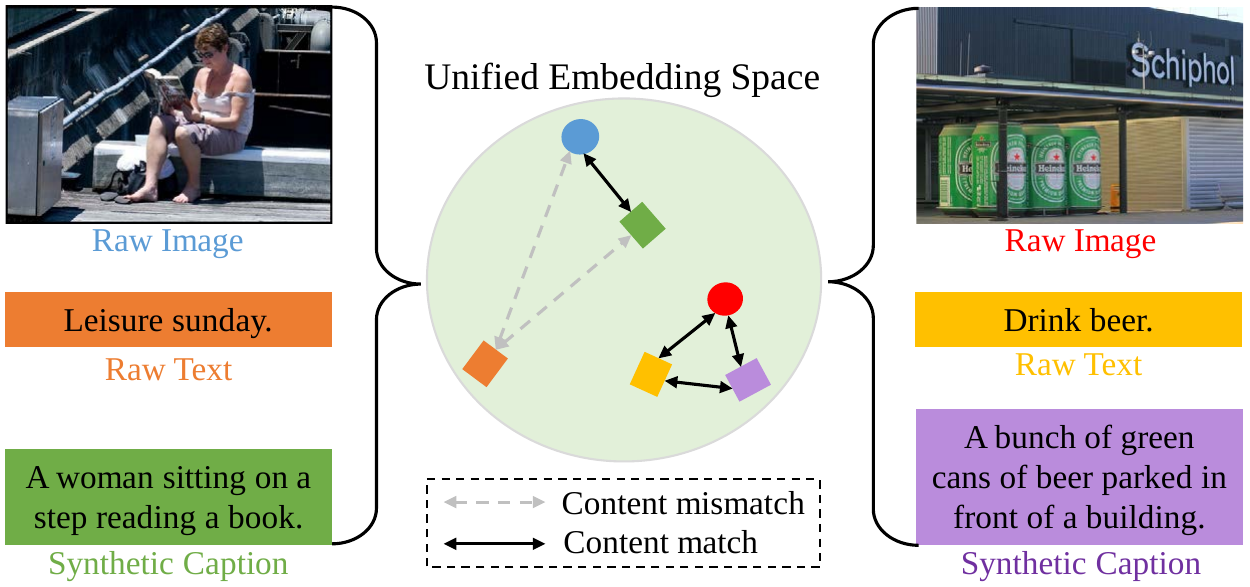} 
\caption{Examples from the YFCC15M dataset to illustrate the mismatched (left) and matched (right) image-text pairs. The synthetic caption is generated from the OFA~\cite{wang2022ofa} model. The raw text description ``Leisure Sunday" is less aligned with the raw image in the left sample, while the synthetic caption ``A woman sitting on a step reading a book" is a more accurate representation. }
\label{fig1}
\vspace{-6mm}
\end{figure}

To facilitate research on large-scale multi-modal models, LAION400M~\cite{schuhmann2021laion} and LAION5B~\cite{schuhmann2022laion} were released, comprising 400 million and 5 billion image-text pairs respectively,  which were filtered using the CLIP model.
However, the current offline filtering approach results in a substantial loss of training data, as the original dataset contains 5 billion image-text pairs. 
Furthermore, this approach may introduce biases due to the limited representation power of the pre-trained model used for filtering.
Despite efforts to curate data for high-quality image-text pairs (\eg, LAION~\cite{schuhmann2021laion,schuhmann2022laion} and YFCC100M~\cite{thomee2016yfcc100m}), 
noisy and poorly-aligned pairs still exist in existing image-text datasets, which can potentially impact the performance of representation learning.

In Fig.~\ref{fig1}, we present two samples from YFCC15M. The raw text of the right sample (``Drink beer'') is correct and matches the content in the image, while the raw text of the left sample (``Leisure Sunday'') is too abstract and does not exactly match the concrete visual signals of the image. 
To alleviate the influence of the noisy and poorly-aligned image-text pairs, BLIP~\cite{li2022blip} bootstraps the captions with an online captioner generating synthetic captions and an online filter removing the noisy ones, while momentum-based methods (\eg, ALBEF~\cite{li2021align} and PSD~\cite{andonian2022robust}) employs soft labels computed using embeddings from a moving average momentum model. However, these filtering and momentum-based methods require additional computation costs and memory consumption.

In this paper, we first employ the OFA~\cite{wang2022ofa} model to generate synthetic captions that are consistent with the image content.
Specifically, the OFA~\cite{wang2022ofa} model is guided by the prompt ``What does the image describe'' to generate synthetic captions. Compared with the raw texts in Fig.~\ref{fig1}, the synthetic captions ``A woman sitting on a step reading a book'' and ``A bunch of green cans of beer parked in front of a building'' provide additional as well as reliable descriptions, such as object information (book, cans), attribute information (green), action information (sitting, parked), and spatial relationship (in front of), which can be used to enhance the representation learning. 

Given the normalized embedding features of the image, raw text, and synthetic caption, 
we propose an Adaptive Language-Image Pre-training (ALIP) method, a bi-path model that integrates raw text supervision and synthetic caption supervision. As the core components of ALIP, the Language Consistency Gate (LCG) and the Description Consistency Gate (DCG) are designed to dynamically adjust the weights of samples and image-text/caption pairs during the training process. The LCG considers the consistency between raw text and synthetic caption to identify the high-quality sample, while the DCG considers the consistency of image-text or image-caption to adjust the contrastive pair loss. 
The adaptive contrastive loss influenced by the above weights substantially reduces the impact of noise data and enhances the efficiency of pre-training data. Extensive experiments show that ALIP achieves state-of-the-art performance on multiple downstream tasks including zero-shot image-text retrieval and linear probe. Experiment results on different model sizes and pre-training datasets also prove the strong robustness of the ALIP. The main contributions of this paper are summarized as follows:
\begin{itemize}[noitemsep,topsep=0pt]
\item We propose a bi-path model that integrates raw text supervision and synthetic caption supervision. Based on the similarity triplet between image, text, and caption, the proposed method can dynamically adjust the weights of samples and image-text/caption pairs through the language consistency gate and description consistency gate.
\item Based on the adaptive weights, we design the adaptive contrastive loss, which can effectively reduce the impact of noise data and improve the pre-training data efficiency.
\item We conduct extensive experiments and prove that ALIP achieves state-of-the-art performance on multiple downstream tasks including zero-shot image-text retrieval and linear probe task.
\end{itemize}

\section{Related Work}

\noindent{\bf Image-Language Pre-training.}
Image-language pre-training aims to improve the performance of downstream vision and language tasks by pre-training the model on large-scale image-text pairs. The milestone work
CLIP~\cite{radford2021learning} has attracted unprecedented attention for its impressive zero-shot recognition ability and excellent transfer ability. Recently, a number of improved methods based on CLIP have been proposed. For more effective training, SLIP~\cite{mu2022slip} significantly improves performance by combining self-supervised learning and CLIP pre-training. DeCLIP~\cite{li2021supervision} explores self-supervision and cross-modal multi-view supervision in the million-scale vision-language pre-training. FILIP~\cite{yao2021filip} learns fine-grained representation for patches in the images and words in the sentences. UniCLIP~\cite{lee2022uniclip} improves data efficiency by integrating contrastive losses defined across multiple domains into a single universal space. HiCLIP~\cite{geng2023hiclip} equips both the visual and language branches in CLIP with hierarchy-aware attention which significantly improves the cross-modal alignment. In this paper, we propose an Adaptive Image-Language Pre-training (ALIP) method to effectively utilize raw text supervision guided by synthetic captions.

\noindent{\bf Noise Alleviation for Contrastive Pre-training.}
Large-scale contrastive pre-training~\cite{radford2021learning,jia2021scaling,schuhmann2021laion,schuhmann2022laion} typically requires dataset sizes of hundreds of millions to billions level. Despite the performance gain obtained by scaling up the training data, the noisy web text is sub-optimal for image-language pre-training. Nevertheless, the cleaning strategy applied to these large-scale data is primitive (\eg, removing samples with short or non-English captions) or biased (\eg, filtering samples based on alignment scores from existing models)~\cite{cherti2022reproducible}. To reduce the adverse effects of noisy image-text pairs in the training data, ALBEF~\cite{li2021align} and PSD~\cite{andonian2022robust} use soft labels computed using embeddings from a moving average momentum model. However, momentum-based approaches are infeasible for large-scale training due to the increased computation and memory consumption. LiT~\cite{zhai2022lit} shows that when a well pre-trained vision encoder is adopted, it is better to lock the vision encoder to protect vision representations from being corrupted by noisy language supervision. However, LiT lacks the ability to align complex text to a fully-trained image encoder, thus underperforming on the multi-modal task, cross-modal retrieval. BLIP~\cite{li2022blip} uses the bootstrapped image-grounded text encoder to filter out noisy captions, but the captioner and filter models need to be finetuned on the COCO dataset beforehand, and these models also increase the overall number of parameters of the model.

\begin{figure*}[t]
\centering
\includegraphics[width=0.95\textwidth]{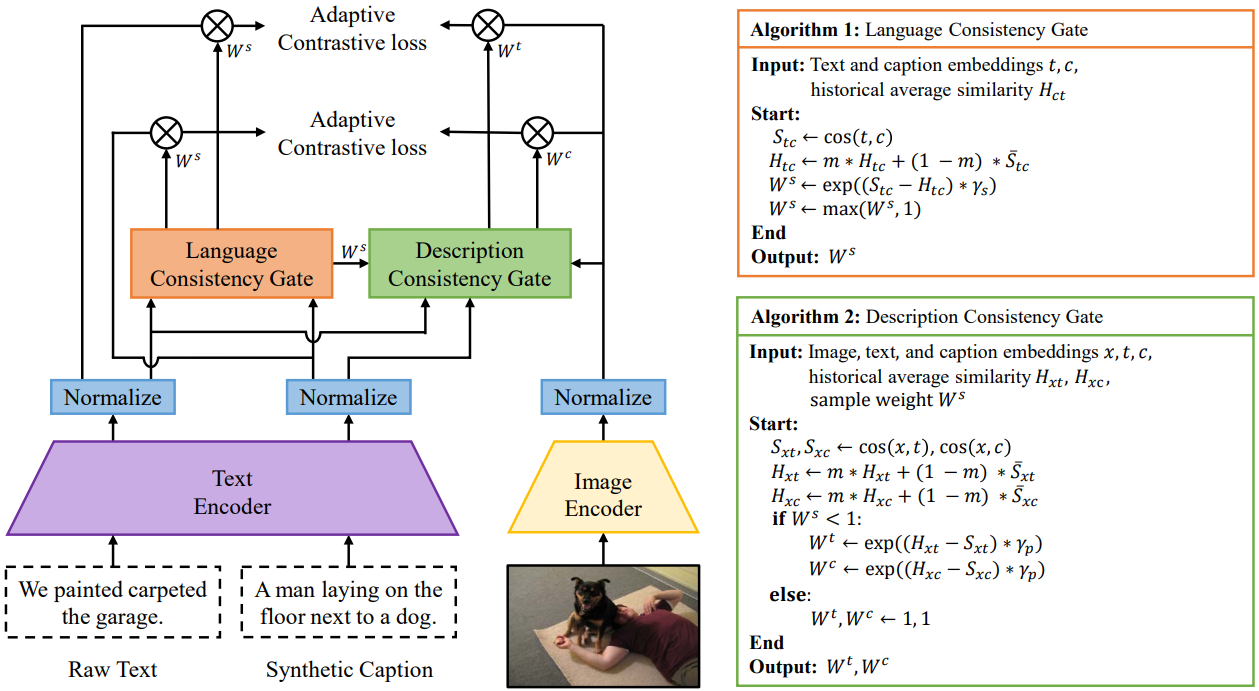}
\caption{
The overall architecture of the proposed Adaptive Language-Image Pre-training (ALIP) method, a bi-path model with triplet input of raw text, synthetic caption, and image. The language consistency gate and description consistency gate are designed to dynamically adjust the weights of samples and image-text/caption pairs during training.
}
\label{fig:architecture}
\vspace{-4mm}
\end{figure*}

In contrast to the previous work, we introduce synthetic captions to alleviate the influence of noise in large-scale vision-language pre-training. Our method can dynamically adjust the weights of samples and image-text pairs through the language consistency gate and description consistency gate. Meanwhile, the adaptive contrastive loss can effectively reduce the impact of noise data and improve the pre-training data efficiency. Our approach is a cheaper alternative since it does not require us to run the expensive online model throughout the training as synthetic captions can be pre-computed and stored offline. In addition, we take advantage of all training samples with adaptive weights instead of directly filtering out image-text pairs.

\section{Methodology}
In this section, we first introduce the model architecture of the proposed method (Sec.~\ref{sec:model}). Then, we delineate the Language Consistency Gate (LCG) and Description Consistency Gate (DCG) in Sec.~\ref{sec:tcg} and~\ref{sec:dcg}.
Lastly, we explain the training objectives of the proposed adaptive contrastive loss for vision-language representation learning (Sec.~\ref{sec:loss}).


\subsection{ALIP Architecture}\label{sec:model}
The primary focus of this paper is on the task of contrastive image-text pre-training. 
Different from the image-text pairs used in CLIP~\cite{radford2021learning}, we adopt the off-the-shelf OFA$_{base}$~\cite{wang2022ofa} model to generate a synthetic caption for each image by applying the prompt  ``What does the image describe?". This method results in a dataset $D = \{(X_i, T_i, C_i)\}_{i=1}^N$ comprising of image-text-caption triplets.
Next, we train a dual encoder model $\Phi = \{\Phi_{\text{image}}, \Phi_{\text{text/caption}}\}$, where $\Phi_{\text{image}}$ represents the image encoder, and $\Phi_{\text{text/caption}}$ denotes the shared text/caption encoder.
We use the shorthand $\mathbf{x} = \Phi_{\text{image}}(X)/\left \| \Phi_{\text{image}}(X) \right \|$, $\mathbf{t} = \Phi_{\text{text}}(T)/\left \| \Phi_{\text{text}}(T) \right \|$, and $\mathbf{c} = \Phi_{\text{caption}}(C)/\left \| \Phi_{\text{caption}}(C) \right \|$ to denote the $l_2$ normalized embeddings of image, text, and caption, respectively, for an image-text-caption triplet $(X, T, C)$.

\begin{figure}
\centering
\includegraphics[width=0.9\linewidth]{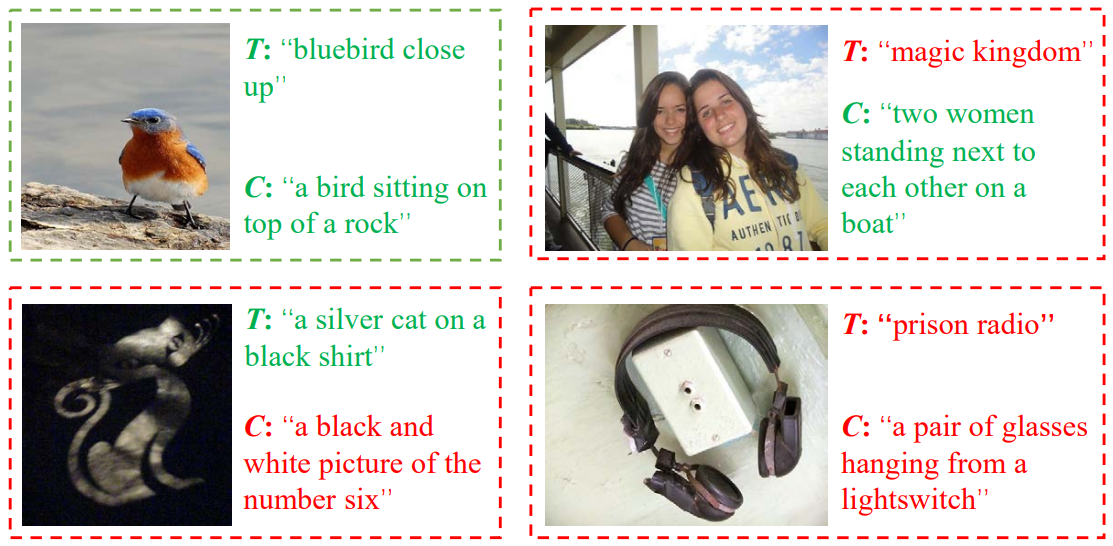}
\caption{Examples of the web raw text $T$ and the synthetic caption $C$ can be categorized into four situations. 
Green and red dotted boxes are used to indicate high-quality and low-quality samples, where green descriptions are considered correct and red descriptions are considered incorrect.}
\vspace{-4mm}
\label{fig:noisecases}
\end{figure}

\begin{figure*}[t]
\centering
\subfigure[$W^s$]{
\label{show_a}
\includegraphics[width=0.3\textwidth]{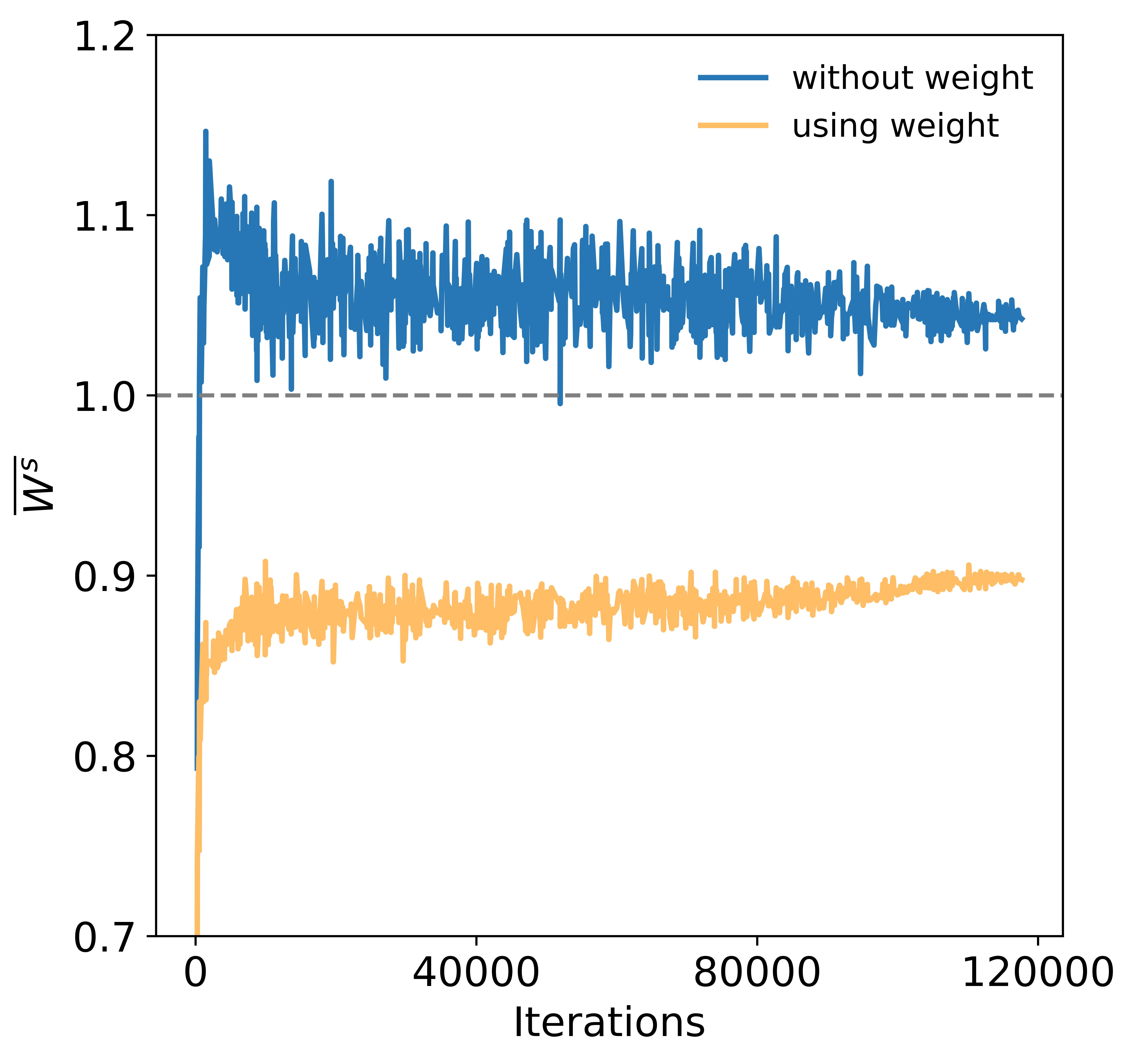}}
\subfigure[$W^t$]{
\label{show_b}
\includegraphics[width=0.3\textwidth]{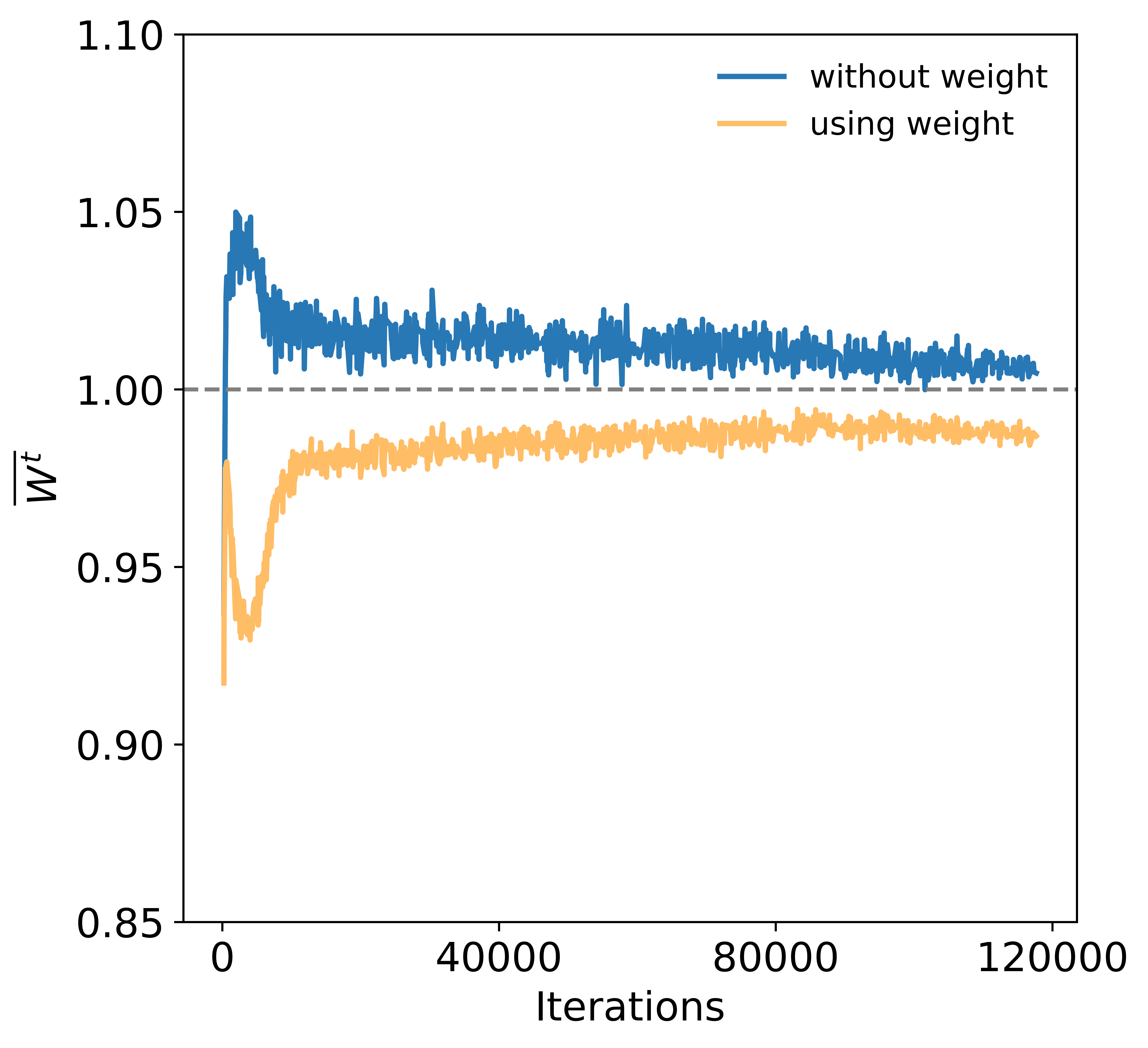}}
\subfigure[$W^c$]{
\label{show_c}
\includegraphics[width=0.3\textwidth]{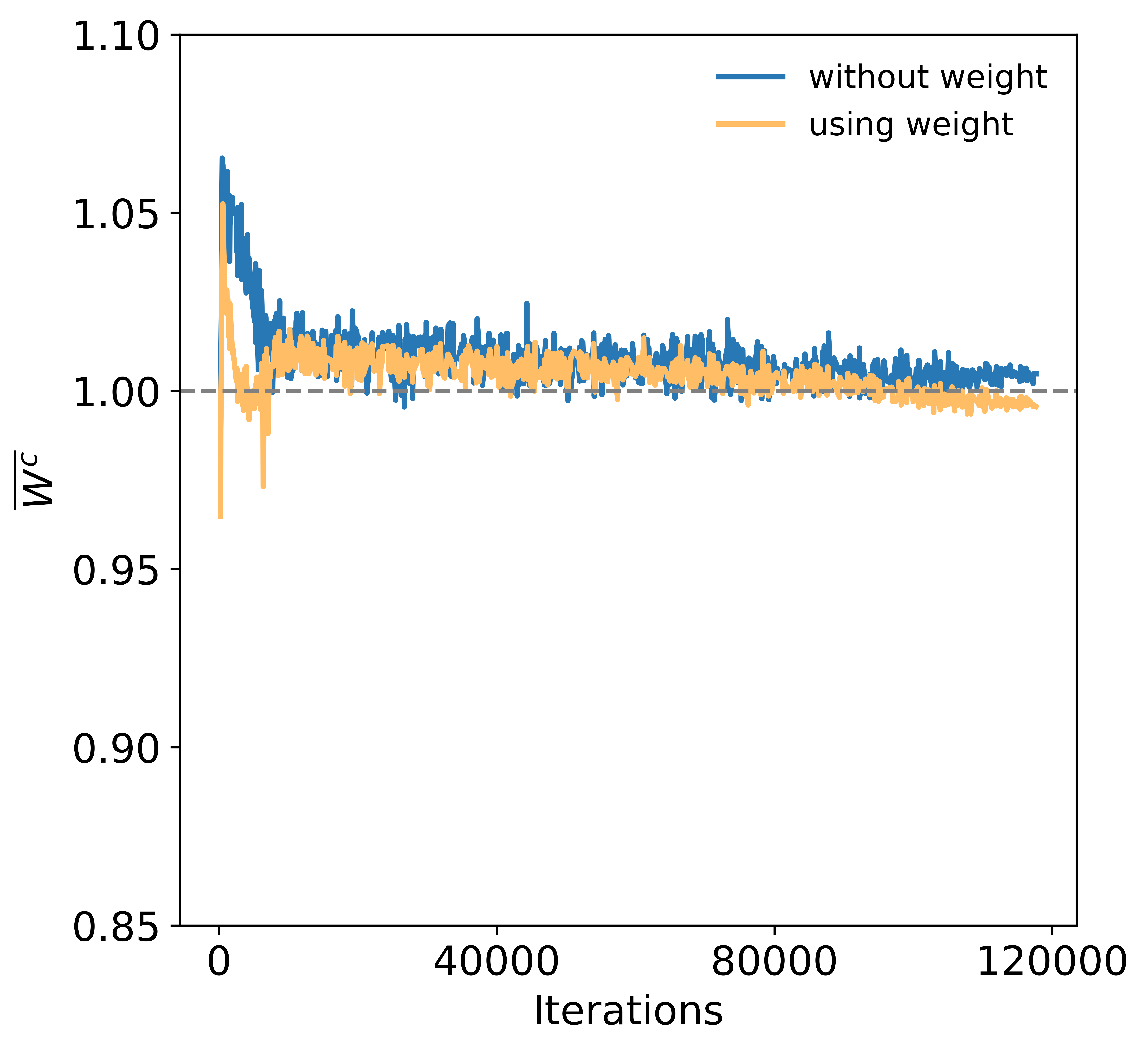}}
\caption{Visualization of different weights during training. Sample weight W$^{s}$ and raw image-text pair weight W$^{t}$ exhibit an obvious decrease. The reduction in image-caption pair weight W$^{c}$ is relatively minor due to the superior consistency in generated captions.}
\vspace{-4mm}
\label{fig:weight}
\end{figure*}

Large vision and language datasets such as YFCC100M \cite{thomee2016yfcc100m} and LAION \cite{li2021supervision} have collected a large number of image-caption pairs from the web, which makes them a good fit for large-scale contrastive pre-training. 
However, these datasets lack semantic-based curation and can contain unilateral or irrelevant raw texts.
Moreover, the automatically generated captions can be also noisy or lack fine granularity\cite{li2022blip}. 
Fig.~\ref{fig:noisecases} shows some examples of the web raw text $T$ and the synthetic caption $C$.
Each box in the figure represents a sample that includes an image along with its corresponding text and caption descriptions.

To mitigate the negative impact of such noise during training, our approach takes into account the similarities between the image, text, and caption triplet.

As illustrated in Fig.~\ref{fig:architecture}, we propose an Adaptive Language-Image Pre-Training (ALIP) approach to make full use of data and reduce the impact of noise. 
Using the $l_2$ normalized triplet embeddings $\mathbf{x}$, $\mathbf{t}$, and $\mathbf{c}$ of image, text, and caption, we can calculate three types of similarities: (1) the similarity between raw text and synthetic caption $S_{tc} = \mathbf{t}*\mathbf{c}$, (2) the similarity between image and raw text $S_{xt} = \mathbf{x}*\mathbf{t}$, and (3) the similarity between synthetic caption and image $S_{xc} = \mathbf{x}*\mathbf{c}$. 
Based on the triplet similarities, we design two gate functions, Language Consistency Gate (LCG) and Description Consistency Gate (DCG). 
More specifically, the LCG predicts a sample weight based on the similarity between raw text embedding and synthetic caption embedding ($S_{tc}$). 
Besides, the DCG computes the image-description weights based on the consistency between the image and text/caption ($S_{xt}$ and $S_{xc}$). Finally, these weights are fed into the adaptive contrastive loss to reduce the impact of noise.

\subsection{Language Consistency Gate}
\label{sec:tcg}

In ensemble learning, confidence in the prediction can be increased when two independent inferences have arrived at the same prediction~\cite{liu2020mesa}. 
Inspired by this, we boost the label confidence of a training sample when the similarity between the raw text and synthetic caption is high. 
To facilitate the accurate assessment of language labels, we introduce a historical average similarity metric ${H}_{tc}$, which is dynamically updated during the training process as follows:
{\small
\begin{equation}
\label{eqn:m_s}
{H}_{tc}=m * {H}_{tc} + (1-m) * \bar{S}_{tc},
\end{equation}
}where $m$ is the momentum and $\bar{S}_{tc}$ denotes the average of $S_{tc}$. 
As both raw text and synthetic caption are explaining the same image, samples with a similarity score $S_{tc}$ higher than the historical average similarity threshold ${H}_{tc}$ will be considered as high-quality samples with reliable language labels. 
By contrast, samples with a lower similarity score are considered as low-quality samples with unreliable language labels. 
To distinguish these two kinds of training samples, we design a sample weight $W^{s}$ and the calculation of $W^{s}$ is given by the following equation: 
{\small 
\begin{equation}
\label{eqn:w_s}
W^{s}= \begin{cases}\mathrm{exp}((S_{tc}-H_{tc}) * \gamma_s), & S_{tc} \leq H_{tc} \\ 1, & S_{tc}> H_{tc}\end{cases},
\end{equation}
}where $\gamma_s$ is a hyper-parameter, and $W^{s}$ is constrained to the range of $\left ( 0,1 \right ]$. 
Consequently, the LCG assigns a lower weight to low-quality samples, reducing the influence of unmatched image-text or image-caption pairs.

\subsection{Description Consistency Gate}
\label{sec:dcg}

While the language consistency gate can identify high-quality training samples, it is important to note that some low-quality samples, as illustrated in Fig.~\ref{fig:noisecases}, can still have well-matched image-text or image-caption pairs that are beneficial for representation learning. 
To fully utilize the pre-training data, we propose the description consistency gate, which computes the image-text pair weight $W^{t}$ and the image-caption pair weight $W^{c}$ for each training image.

Due to the considerable discrepancy between raw text and synthetic caption (which is discussed in Sec.~\ref{section:ablation}), we separately record the historical image-text pair similarity ${H}_{xt}$ and the historical image-caption pair similarity ${H}_{xc}$, which are updated dynamically as follows:
{\small 
\begin{equation}
\label{eqn:his_p}
\begin{split}
{H}_{x t}=m * {H}_{xt} + (1-m) * \bar{S}_{xt}, \\
{H}_{x c}=m * {H}_{xc} + (1-m) * \bar{S}_{xc},
\end{split}
\end{equation}
}where the $\bar{S}_{xt}$ and $\bar{S}_{xc}$ denote the average similarity of the image-text and image-caption pairs. 
Based on the similarity scores and historical image-text or image-caption pair similarity, the description consistency gate computes the pair weights $W^{t}$ and $W^{c}$. 
\begin{small} 
\begin{equation}
\label{eqn:w_p}
\begin{split}
W^{t}= \begin{cases}\mathrm{exp}((S_{x t}-H_{x t}) * \gamma_p), & W^{s} < 1 \\ 1, & W^{s}= 1\end{cases} \\
W^{c}= \begin{cases}\mathrm{exp}((S_{x c}-H_{x c}) * \gamma_p), & W^{s} < 1 \\ 1, & W^{s}= 1\end{cases}
\end{split}
\end{equation}
\end{small}
The pair weight $W^{t}$ and $W^{c}$ share a common hyper-parameter $\gamma_p$. 
When $W^{s}=1$, the training sample is considered to be high-quality and both $W^{t}$ and $W^{c}$ are set to 1. However, when $W^{s}<1$, $W^{t}$ will be larger than $1$ if 
$S_{x t}>H_{x t}$, and $W^{c}$ will be larger than $1$ if $S_{x c}>H_{x c}$. 
One noteworthy benefit of introducing $W^{t}$ and $W^{c}$ is that they are capable of precisely exploiting high-quality image-text or image-caption pairs from low-quality samples.


\subsection{Adaptive Contrastive Loss}
\label{sec:loss}

CLIP \cite{radford2021learning} utilizes the InfoNCE loss~\cite{oord2018representation} for multi-modal alignment. 
Given a mini-batch $D = \{(x_i,t_i)\}_{i=1}^N$ of image-text feature embeddings, the multi-modal InfoNCE objective is defined as, 
{\small
\begin{equation}
\label{eqn:nce}
L_{\text{NCE}} = -\sum_{i=1}^N\left[\log\frac{e^{x_i^\top t_i/\tau}}{\sum_{j}e^{x_i^\top t_j/\tau}}+\log\frac{e^{ x_i^\top t_i/\tau}}{\sum_{j}e^{ x_j^\top t_i/\tau}}\right],
\end{equation}
}where $\tau$ is the temperature parameter. Even though the InfoNCE loss has achieved huge success in language-vision pre-training,
when learning from large-scale noisy web data, the uniform weighting of all training samples can lead to adverse effects on representation learning.

In this paper, we propose an adaptive contrastive loss that incorporates additional sample weight and pair weight into the InfoNCE loss. 
By dynamically adjusting the sample and pair weights during the training process, the adaptive loss can significantly reduce the impact of noise. Specifically, given a mini-batch $D = \{(x_i,t_i,c_i)\}_{i=1}^N$ of image-text-caption feature embeddings, 
the adaptive contrastive loss $L_{xt}$ and $L_{xc}$ between the image-text pair and image-caption pair are defined by the following formula:

{\footnotesize
\begin{equation}
\label{eqn:CLIP}
\begin{split}
L_{\text{xt}} = -\sum_{i=1}^N W^s_i W^{t}_i\left[\log\frac{e^{x_i^\top t_i/\tau}}{\sum_{j}e^{x_i^\top t_j/\tau}}+\log\frac{ e^{ x_i^\top t_i/\tau}}{\sum_{j}e^{ x_j^\top t_i/\tau}}\right], \\
L_{\text{xc}} = -\sum_{i=1}^N W^s_i W^{c}_i\left[\log\frac{e^{x_i^\top c_i/\tau}}{\sum_{j}e^{x_i^\top c_j/\tau}}+\log\frac{e^{ x_i^\top c_i/\tau}}{\sum_{j}e^{ x_j^\top c_i/\tau}}\right],
\end{split}
\end{equation}
}where $W^s_i$ is calculated by the language consistency gate, and $W^t_i$ and $W^c_i$ are computed by the description consistency gate. Finally, the overall loss function of our ALIP is defined by combining the bi-path contrastive loss $L_{ALIP}=L_{xt} + L_{xc}$. Fig.~\ref{fig:weight} illustrates the variation of weights during the training process. ALIP is capable of effectively adjusting the weights to mitigate the impact of noise.

\begin{table*}[t]
    \caption{Zero-shot image-text retrieval on the test splits of Flickr30k and MSCOCO. All models are pre-trained on YFCC15M, and ALIP creates new state-of-the-art results on all the metrics.}
    \vspace{1mm}
    \label{tab:retrieval}
    \centering
    \resizebox{0.98\linewidth}{!}{
        \centering
        \begin{sc}
        \begin{tabular}{lcccccccccccc}
        \toprule
        & \multicolumn{6}{c}{Text retrieval} & \multicolumn{6}{c}{Image retrieval} \\
        & \multicolumn{3}{c}{Flickr30k} & \multicolumn{3}{c}{MSCOCO} & \multicolumn{3}{c}{Flickr30k} & \multicolumn{3}{c}{MSCOCO} \\
        Method &  R@1 & R@5 & R@10 & R@1 & R@5 & R@10 & R@1 & R@5 & R@10 & R@1 & R@5 & R@10 \\
        \midrule
        CLIP-ViT-B/32\cite{radford2021learning} & 34.9 & 63.9 & 75.9 & 20.8 & 43.9 & 55.7 & 23.4 & 47.2 & 58.9 & 13.0 & 31.7 & 42.7 \\
        SLIP-ViT-B/32~\cite{mu2022slip} & 47.8 & 76.5 & 85.9 & 27.7 & 52.6 & 63.9 & 32.3 & 58.7 & 68.8 & 18.2 & 39.2 & 51.0 \\
        DeCLIP-ViT-B/32~\cite{li2021supervision} & 51.4 & 80.2 & 88.9 & 28.3 & 53.2 & 64.5 & 34.3 & 60.3 & 70.7 & 18.4 & 39.6 & 51.4 \\
        UniCLIP-ViT-B/32~\cite{lee2022uniclip} & 52.3 & 81.6 & 89.0 & 32.0 & 57.7 & 69.2 & 34.8 & 62.0 & 72.0 & 20.2 & 43.2 & 54.4 \\
        HiCLIP-ViT-B/32~\cite{geng2023hiclip} & - & - & - & 34.2 & 60.3 & 70.9 & - & - & - & 20.6 & 43.8 & 55.3 \\
        HiDeCLIP-ViT-B/32~\cite{geng2023hiclip} & - & - & - & 38.7 & 64.4 & 74.8 & - & - & - & 23.9 & 48.2 & 60.1 \\
        \midrule
        ALIP-ViT-B/32 & \textbf{70.5} & \textbf{91.9} & \textbf{95.7} & \textbf{46.8} & \textbf{72.4} & \textbf{81.8} & \textbf{48.9} & \textbf{75.1} & \textbf{82.9} & \textbf{29.3} & \textbf{54.4} & \textbf{65.4} \\
        \bottomrule
        \end{tabular}
\end{sc}}
\end{table*}

\begin{table*}[t]
\caption{Linear probe performance on 10 downstream datasets. ALIP achieves higher average accuracy with an improvement 
 of 1.4$\sim$9.2\%.}
\vspace{1mm}
\label{tab:linear}
\centering
\resizebox{0.98\linewidth}{!}{
    \begin{sc}
    \begin{tabular}{lcccccccccccccc}
        \toprule
        Method & \shortstack{Pre-train \\ dataset} &  \rotatebox[origin=lb]{90}{\smash{CIFAR10}} & \rotatebox[origin=lb]{90}{\smash{CIFAR100}} &  \rotatebox[origin=lb]{90}{\smash{Food101}} & \rotatebox[origin=lb]{90}{\smash{Pets}} &  \rotatebox[origin=lb]{90}{\smash{Flowers}} & 
        \rotatebox[origin=lb]{90}{\smash{SUN397}} &
        \rotatebox[origin=lb]{90}{\smash{Cars}} & 
        \rotatebox[origin=lb]{90}{\smash{DTD}} & 
        \rotatebox[origin=lb]{90}{\smash{Caltech101}} & 
        \rotatebox[origin=lb]{90}{\smash{Aircraft}} & 
        \rotatebox[origin=lb]{90}{\smash{Average}}  \\
        \midrule

        CLIP-ViT-B/32\cite{radford2021learning} & YFCC15M & 86.5 & 64.7 & 69.2 & 64.6 & 90.6 & 66.0 & 24.9 & 61.3 & 79.1 & 23.1 & 63.0 \\
        DeCLIP-ViT-B/32~\cite{li2021supervision} & YFCC15M & 89.2 & 69.0 & 75.4 & 72.2 & 94.4 & 71.6 & 31.0 & 68.8 & 87.9 & 27.6 & 68.7 \\
        HiCLIP-ViT-B/32~\cite{geng2023hiclip}  & YFCC15M & 89.5 & 71.1 & 73.5 & 70.6 & 91.9 & 68.8 & 30.8 & 63.9 & 84.8 & 27.4 & 67.2 \\
        HiDeCLIP-ViT-B/32~\cite{geng2023hiclip} & YFCC15M & 88.1 & 70.7 & \textbf{77.6} & 75.5 & \textbf{95.6} & 72.2 & \textbf{36.0} & 70.1 & \textbf{90.0} & 32.6 & 70.8 \\
        \midrule
        ALIP-ViT-B/32 & YFCC15M & \textbf{94.3} & \textbf{77.8} & 75.8 & \textbf{76.0} & 95.1 & \textbf{73.3} & 33.6 & \textbf{71.7} & 88.5 & \textbf{36.1} & \textbf{72.2}  \\
        \bottomrule
  \end{tabular}
  \end{sc}} 
\end{table*}

\begin{table*}[h]
\caption{ zero-shot classification performance on 11 downstream datasets.}
\vspace{1mm}
\label{tab:transfer}
\centering
 
\resizebox{0.98\linewidth}{!}{
    \begin{sc}
    \begin{tabular}{lccccccccccccccc}
        \toprule
        
        Method & \shortstack{Pre-train \\ dataset} &  \rotatebox[origin=lb]{90}{\smash{CIFAR10}} & \rotatebox[origin=lb]{90}{\smash{CIFAR100}} &  \rotatebox[origin=lb]{90}{\smash{Food101}} & \rotatebox[origin=lb]{90}{\smash{Pets}} &  \rotatebox[origin=lb]{90}{\smash{Flowers}} & 
        \rotatebox[origin=lb]{90}{\smash{SUN397}} &
        \rotatebox[origin=lb]{90}{\smash{Cars}} & 
        \rotatebox[origin=lb]{90}{\smash{DTD}} & 
        \rotatebox[origin=lb]{90}{\smash{Caltech101}} & 
        \rotatebox[origin=lb]{90}{\smash{Aircraft}} & 
        \rotatebox[origin=lb]{90}{\smash{ImageNet}}  &
        \rotatebox[origin=lb]{90}{\smash{Average}}  \\
        \midrule

        CLIP-ViT-B/32\cite{radford2021learning} & YFCC15M & 63.7 & 33.2 & 34.6 & 20.1 & 50.1 & 35.7 & 2.6 & 15.5 & 59.9 & 1.2 & 32.8 & 31.8 \\

        SLIP-ViT-B/32~\cite{mu2022slip} & YFCC15M & 50.7 & 25.5 & 33.3 & 23.5 & 49.0 & 34.7 & 2.8 & 14.4 & 59.9 & 1.7 & 34.3 & 30.0 \\
        
        FILIP-ViT-B/32~\cite{yao2021filip} & YFCC15M & 65.5 & 33.5 & 43.1 & 24.1 & 52.7 & 50.7 & 3.3 & 24.3 & 68.8 & 3.2 & 39.5 & 37.2 \\
        
        DeCLIP-ViT-B/32~\cite{li2021supervision} & YFCC15M & 66.7 & 38.7 & 52.5 & 33.8 & 60.8 & 50.3 & 3.8 & 27.7 & 74.7 & 2.1 & 43.2 & 41.3 \\
        
        DeFILIP-ViT-B/32~\cite{cui2022democratizing} & YFCC15M & 70.1 & 46.8 & 54.5 & 40.3 & 63.7 & 52.4 & 4.6 & 30.2 & 75.0 & 3.3 & 45.0 & 44.2 \\
        HiCLIP-ViT-B/32~\cite{geng2023hiclip} & YFCC15M & 74.1 & 46.0 & 51.2 & 37.8 & 60.9 & 50.6 & 4.5 & 23.1 & 67.4 & 3.6 & 40.5 & 41.8 \\
        HiDeCLIP-ViT-B/32~\cite{geng2023hiclip} & YFCC15M & 65.1 & 39.4 & \textbf{56.3} & \textbf{43.6} & \textbf{64.1} & \textbf{55.4} & \textbf{5.4} & \textbf{34.0} & \textbf{77.0} & \textbf{4.6} & \textbf{45.9} & \textbf{44.6} \\
        \midrule
        ALIP-ViT-B/32 & YFCC15M & \textbf{83.8} & \textbf{51.9} & 45.4 & 30.7 & 54.8 & 47.8 & 3.4 & 23.2 & 74.1 & 2.7 & 40.3 & 41.7  \\
        
        \bottomrule
  \end{tabular}
  \end{sc}} 
\end{table*}

\section{Experiments}
\subsection{Experimental Settings}
\noindent{\bf Pre-training Datasets.} We train our model on the YFCC15M dataset, which is a subset of YFCC100M~\cite{thomee2016yfcc100m} filtered by DeCLIP~\cite{li2021supervision}. To further verify the effectiveness and generalizability of ALIP, we randomly select subsets of 10M and 30M from the LAION400M dataset~\cite{schuhmann2021laion} and conduct a series of experiments with different model sizes and pre-training data scales.

\noindent{\bf Downstream Datasets.} Following recent works~\cite{mu2022slip,yao2021filip,lee2022uniclip}, we evaluate the effectiveness of our approach in zero-shot image-text retrieval tasks on the Flickr30K~\cite{plummer2015flickr30k} and MSCOCO~\cite{pont2020connecting} benchmarks. Besides, consistent with HiCLIP~\cite{geng2023hiclip}, we report the linear probe performance over 10 datasets and the zero-shot classification performance over 11 datasets, including CIFAR10 \& CIFAR100~\cite{krizhevsky2009learning}, Food101~\cite{bossard2014food}, Oxford Pets~\cite{parkhi2012cats}, Flowers102~\cite{nilsback2008automated}, SUN397~\cite{xiao2010sun}, Stanford Cars~\cite{KrauseStarkDengFei-Fei_3DRR2013}, DTD~\cite{cimpoi2014describing}, Caltech101~\cite{fei2004learning}, FGVC Aircraft~\cite{maji2013fine}, and ImageNet~\cite{deng2009imagenet}.

\noindent{\bf Implementation Details.}
We employ OFA$_{base}$ to generate synthetic captions. The image encoder and text encoder in ALIP follow the same architecture as in CLIP~\cite{radford2021learning}. We use AdamW~\cite{loshchilov2017decoupled} as the optimizer with an initial learning rate of $0.001$ and a weight decay of $0.2$. Consistent with CLIP, we set $\beta_{1}$ to $0.9$ and $\beta_{2}$ to $0.98$ to improve training stability. The input image size is $224 \times 224$, and the input text sequence length is truncated or padded to $77$. The temperature parameter $\tau$ is initialized to $0.07$. To ensure a fair comparison with baselines, we train ALIP for 32 epochs with a batch size of $4096$ on $16$ NVIDIA V100 GPUs.

\subsection{Experimental Results}
We compare ALIP with state-of-the-art approaches by using YFCC15M. Following HiCLIP~\cite{geng2023hiclip}, We report the performance of ALIP in zero-shot text-image retrieval, linear probe, and zero-shot classification, respectively. 

\noindent{\bf Zero-shot Image-text Retrieval.}
In Tab.~\ref{tab:retrieval}, we present a comparison of our method with state-of-the-art approaches in zero-shot image-text retrieval on Flickr30k and MSCOCO. Our proposed ALIP achieves new state-of-the-art results on all evaluation metrics. Specifically, ALIP achieves 46.8\%/29.3\% I2T/T2I retrieval Recall@1 on MSCOCO, which is 8.1\%/5.4\% higher than HiDeCLIP and 12.6\%/8.7\% higher than HiCLIP. Similarly, ALIP demonstrates significant improvements of 18.2\% to 35.6\% and 14.1\% to 25.5\% on Flickr30K. The performance improvement is mainly attributed to the more robust image description supervision 
as ALIP can dynamically adjust the weights of samples and image-text/caption pairs to reduce the impact of noise.

\noindent{\bf Linear Probe.}
Following the same evaluation setting as CLIP, we freeze the ALIP model and only train a logistic regression classifier. 
In Tab.~\ref{tab:linear}, we report our linear probe performance on 10 downstream datasets by referring to HiCLIP~\cite{geng2023hiclip}. Compared with the baseline methods, our ALIP yields an improvement of 1.4\% to 9.2\% on average, and it surpasses HiCLIP on all datasets and HiDeCLIP on 5 out of 10 datasets. Although ALIP does not exhibit superior performance to HiDeCLIP in half of the datasets, the performance gaps are marginal. Remarkably, compared with HiDeCLIP, ALIP observes a significant performance boost of 6.2\%, 7.1\%, and 3.5\% on the CIFAR10, CIFAR100, and Aircraft datasets, respectively. The performance improvement demonstrates that ALIP can effectively enhance the representation power in instance discrimination.

\noindent{\bf Zero-shot Classification.}
We also present our performance on 11 zero-shot classification benchmarks. The prompt templates and class names used for evaluation are consistent with HiCLIP~\cite{geng2023hiclip} and SLIP~\cite{mu2022slip}. As shown in Tab~\ref{tab:transfer}, ALIP achieves substantial improvement only on CIFAR10 and CIFAR100, but it still lags behind HiDeCLIP in terms of zero-shot accuracy. This performance gap is mainly due to the coarse-grained synthetic captions generated by the OFA model. For instance, the OFA model can only recognize the presence of flowers in an image and can not identify the specific species of flower. 
Besides, ALIP aims to reduce the impact of noisy image-text pairs, and it does not fully account for the hierarchical nature
of fine-grained semantics, as does HiDeCLIP.

\begin{table}[t]
\caption{Ablation on the sample weight $W^{s}$ and image-text/caption pair weights $W^{t}$ and $W^{c}$. All models are pre-trained on YFCC15M.}
\vspace{1mm}
\label{tab:ablationonweight}
\centering
\resizebox{1.0\columnwidth}{!}{
\begin{sc}
\begin{tabular}{lccccccc}
\toprule
& \multicolumn{3}{c}{Weight}& \multicolumn{2}{c}{Flickr30K} & \multicolumn{2}{c}{MSCOCO} \\ 
Methods  & $W^{s}$ & $W^{t}$& $W^{c}$& I2T    & T2I   & I2T   & T2I   \\ \midrule
ALIP-ViT-B/32 & $\times$    &   $\times$ &   $\times$   & 68.7  & 48.1  & 45.1  & 27.9  \\
ALIP-ViT-B/32 & \checkmark &    $\times$ &   $\times$  & 69.8  & 49.1  & 45.8  & 29.1  \\
ALIP-ViT-B/32 & $\times$    &   \checkmark &   $\times$ & 69.5  & \textbf{49.4}  & 45.6  & 29.3  \\
ALIP-ViT-B/32 & $\times$    &   $\times$ &  \checkmark  & 68.9  & 48.3  & 45.3  & 28.7  \\
ALIP-ViT-B/32 &  \checkmark &   \checkmark &   \checkmark & \textbf{70.5}  & 48.9  & \textbf{46.8}  & \textbf{29.3}  \\
\bottomrule
\end{tabular}
\end{sc}
}
\end{table}

\begin{figure}
\centering
\subfigure[Zero-shot text retrieval]{
\includegraphics[height=0.21\textwidth]{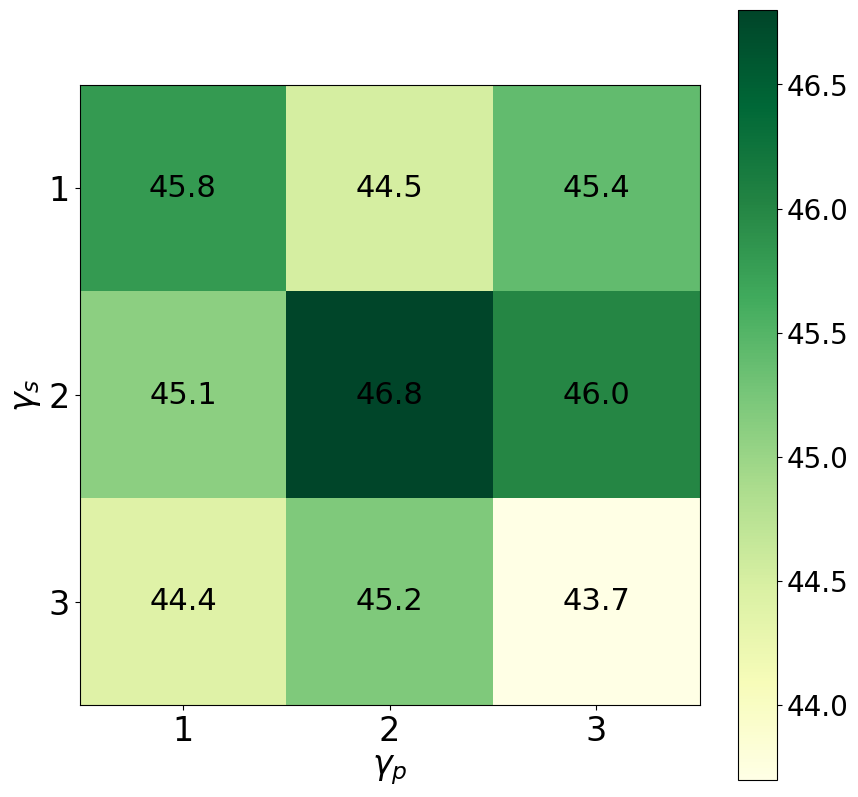}}
\subfigure[Zero-shot classification]{
\includegraphics[height=0.21\textwidth]{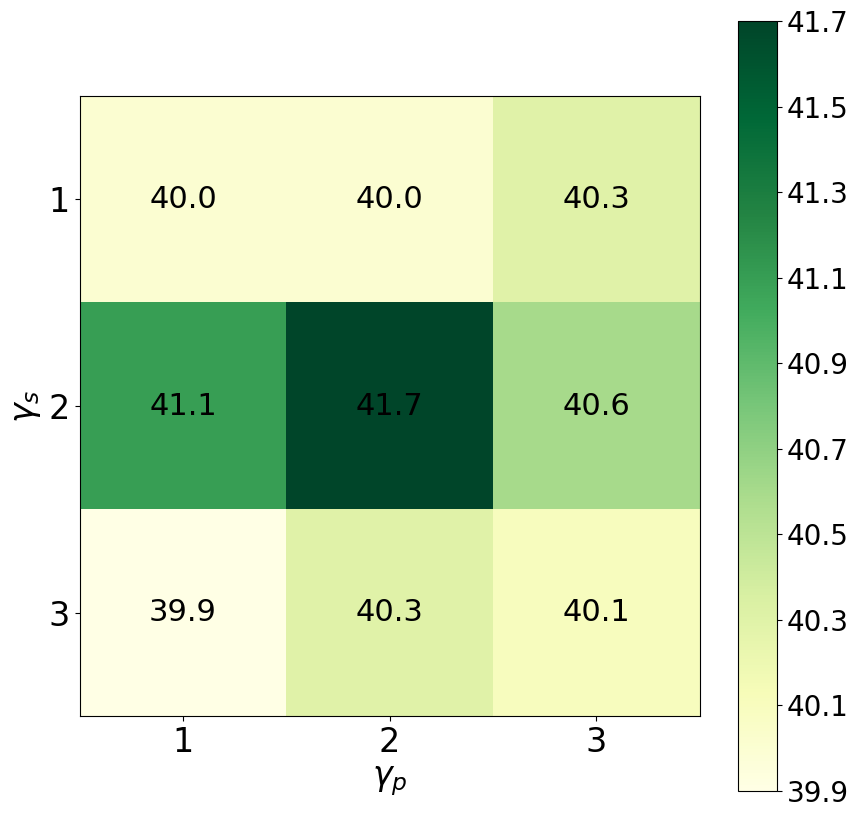}}
\caption{Ablation on the parameters $\gamma_{s}$ and $\gamma_{p}$. (a) is the zero-shot text retrieval Recall@1 on MSCOCO. (b) is the average zero-shot classification accuracy on 11 datasets.}
\label{fig:metrix}
\vspace{-4mm}
\end{figure}

\begin{table}[h]
\caption{The influence of the caption model in the linear probe and zero-shot classification tasks.}
\vspace{1mm}
\label{ablation:ofasize}
\centering
    \resizebox{1.0\columnwidth}{!}{
    \begin{sc}
    \begin{tabular}{lcccc}
    \toprule
    Method & \shortstack{Caption\\ Model} & \shortstack{Linear probe\\ avg} & \shortstack{Zero-shot \\ avg} \\
    \midrule
    ALIP-ViT-B/32 & OFA$_{base}$  & 72.2  & 41.7 \\
    ALIP-ViT-B/32 & OFA$_{large}$  & 72.3  & 42.0 \\
     \bottomrule
    \end{tabular} 
    \end{sc}}
\vspace{-4mm}
\end{table}

\begin{table*}[t]
  \caption{The linear probe and zero-shot classification performance of CLIP-ViT-B/32 trained on LAION10M.}
  \vspace{1mm}
  \label{ablation:synthetic}
    \centering
    \resizebox{1.0\linewidth}{!}{
        \begin{sc}
        \begin{tabular}{lcccccccccccccc}
        \toprule
        Method & \shortstack{Pre-train\\ data} &  \rotatebox[origin=lb]{90}{\smash{CIFAR10}} & \rotatebox[origin=lb]{90}{\smash{CIFAR100}} &  \rotatebox[origin=lb]{90}{\smash{Food101}} & \rotatebox[origin=lb]{90}{\smash{Pets}} &  \rotatebox[origin=lb]{90}{\smash{Flowers}} & 
        \rotatebox[origin=lb]{90}{\smash{SUN397}} &
        \rotatebox[origin=lb]{90}{\smash{Cars}} & 
        \rotatebox[origin=lb]{90}{\smash{DTD}} & 
        \rotatebox[origin=lb]{90}{\smash{Caltech101}} & 
        \rotatebox[origin=lb]{90}{\smash{Aircraft}} & 
        \rotatebox[origin=lb]{90}{\smash{Average}}  \\
        \midrule
        \small\textit{Linear probe:}\\
        CLIP-ViT-B/32 & \it Raw text-image & 91.2 & 74.8 & 66.9 & 71.0 & 63.0 & 89.5 & 71.1 & 68.5 & 40.3 & 84.7 & 72.1 \\
        CLIP-ViT-B/32 & \it Synthetic caption-image & 90.7 & 71.9 & 65.1 & 68.6 & 63.8 & 88.2 & 39.5 & 68.3 & 40.3 & 85.5 & 68.2 \\
        \midrule
        
        \small\textit{zero-shot classification:}\\
        CLIP-ViT-B/32 & \it Raw text-image & 78.5 & 49.3 & 42.0 & 42.5 & 28.6 & 40.8 & 39.9 & 23.7 & 73.2 & 4.4 & 42.3 \\
        CLIP-ViT-B/32 & \it Synthetic caption-image & 57.1 & 21.1 & 9.9 & 8.3 & 4.8 & 10.8 & 2.8 & 9.2 & 39.5 & 1.0 & 16.5 \\
       
        \bottomrule
    \end{tabular}
\end{sc}}
\vspace{-4mm}
\end{table*}

\begin{figure*}
\centering
    \subfigure[]{
    \label{b32_10m}
    \includegraphics[width=0.32\textwidth]{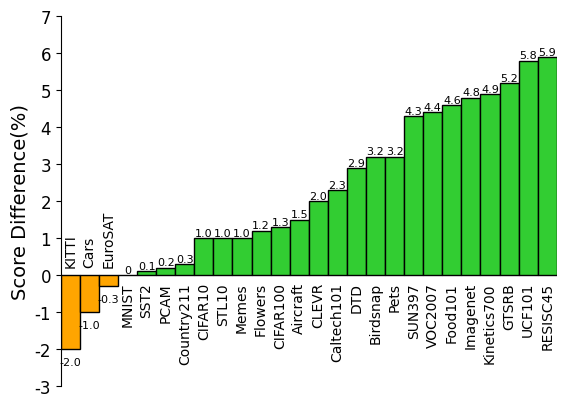}}
    \subfigure[]{
    \label{b16_10m}
    \includegraphics[width=0.32\textwidth]{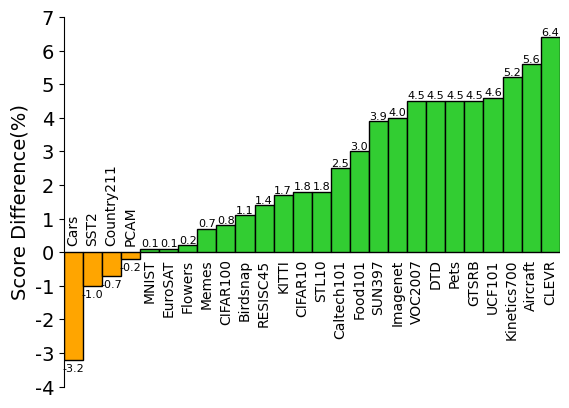}}
    \subfigure[]{
    \label{b32_30m}
    \includegraphics[width=0.32\textwidth]{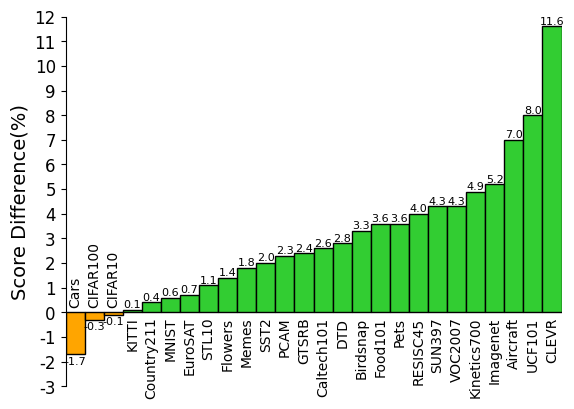}}
    \caption{Linear probe performance comparison between ALIP and CLIP on 27 downstream datasets. (a) ALIP-ViT-B/32 vs. CLIP-ViT-B/32 on LAION10M; (b) ALIP-ViT-B/16 vs. CLIP-ViT-B/16 on LAION10M; (c) ALIP-ViT-B/32 vs. CLIP-ViT-B/32 on LAION30M.}
\label{fig:laion}
\vspace{-5mm}
\end{figure*}

\subsection{Ablation Study}
\label{section:ablation}
\noindent{\bf Ablation on Adaptive Weights.}
To further explore the effectiveness of the sample weight and image-text/caption pair weight, we perform ablation experiments based on the zero-shot image-text retrieval task. The retrieval Recall@1 for I2T/T2I on Flickr30K and MSCOCO is presented in Tab.~\ref{tab:ablationonweight}. Our results indicate that both the sample weight and image-text pair weight improved retrieval performance. Additionally, the improvements are more significant when applying $W^{t}$ alone than applying $W^{c}$, indicating that raw texts have weaker description consistency. 

\noindent{\bf Ablation on the Parameters $\gamma_{s}$ and $\gamma_{p}$.}
The parameters $\gamma_{s}$ and $\gamma_{p}$ directly affect the sample weight and pair weight. In Fig.~\ref{fig:metrix}, we show the zero-shot text retrieval Recall@1 on MSCOCO and the average zero-shot classification accuracy on 11 downstream datasets under different parameter settings. When $\gamma_{s} = 2$ and $\gamma_{p} = 2$, ALIP achieves the best performance on both zero-shot text retrieval and zero-shot classification tasks.

\noindent{\bf Ablation on Different Capacity of Caption Model.}
Given the significance of synthetic captions in this study, we investigate the impact of captions generated by different sizes of the OFA model on downstream tasks. Specifically, in addition to OFA$_{base}$, we employ the OFA$_{large}$ model which has 470M parameters to generate synthetic captions on the YFCC15M dataset. Then, we train the ALIP-ViT-B/32 and evaluate the average accuracy of linear probe and zero-shot classification on 10 and 11 datasets. The experiment results are presented in Tab.~\ref{ablation:ofasize}, it is worth noting that despite the OFA$_{large}$ model having 2.5 times the parameters of OFA$_{base}$, it only yields a marginal improvement 0.1\% on linear probe and 0.4\% on zero-shot classification. Additionally, we provide some examples of synthetic captions generated by OFA$_{base}$ and OFA$_{large}$ in the supplementary material. While the synthetic captions generated by OFA$_{large}$ are of higher quality, they still remain coarse-grained descriptions.

\noindent{\bf Analysis of Raw Text and Synthetic Caption.}

\begin{figure}
\centering
\subfigure[Distribution of similarity]{
\includegraphics[height=0.18\textwidth]{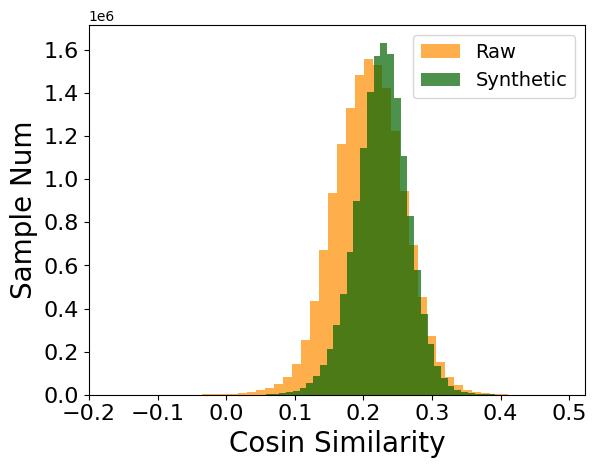}}
\subfigure[Distribution of token num]{
\includegraphics[height=0.18\textwidth]{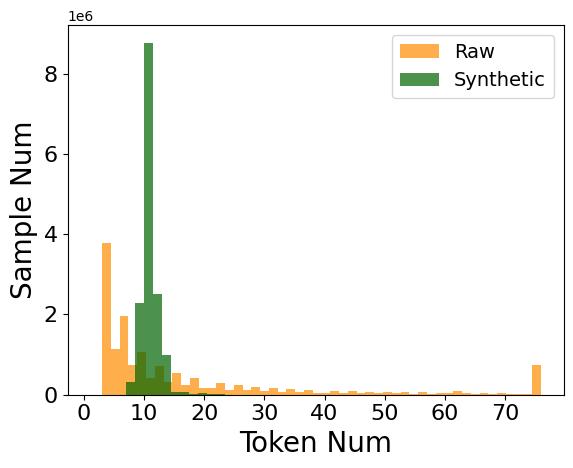}}
\caption{We conduct a statistical analysis of raw text and synthetic caption on YFCC15M. (a) is the image-text/caption similarity distribution; (b) is the token number distribution of the raw texts and synthetic captions.}
\vspace{-5mm}
\label{fig5}
\end{figure}

To examine the distinctions between synthetic captions and raw texts, we conduct a statistical analysis of the token counts and use the CLIP ViT-L/14 model to compute the distribution of similarity between raw and synthetic image-text pairs. As illustrated in Fig.~\ref{fig5}, in comparison to raw texts, synthetic captions demonstrate a higher average similarity and more compact similarity distribution. Additionally, we observe that the number of tokens in the synthetic caption is predominantly concentrated between 10 and 15, which is significantly lower than in raw text. 

To better investigate the performance disparities between synthetic caption and raw text in downstream tasks, based on LAION10M, we train CLIP-B/32 on raw and synthetic image-text pairs respectively. We present the linear probe and zero-shot classification performance in Tab.~\ref{ablation:synthetic}. Compared with the CLIP-B/32 trained on the raw image-text pairs, the CLIP-B/32 trained on the synthetic caption-image pairs achieves similar or better linear probe performance on all the datasets except Cars. However, the zero-shot results reveal a significant deficiency of synthetic captions in zero-shot classification task. This is mainly due to the coarse granularity of the synthetic captions, which also explains the inferior performance of ALIP in Tab.~\ref{tab:transfer}.

\begin{figure}
\includegraphics[width=1.0 \linewidth]{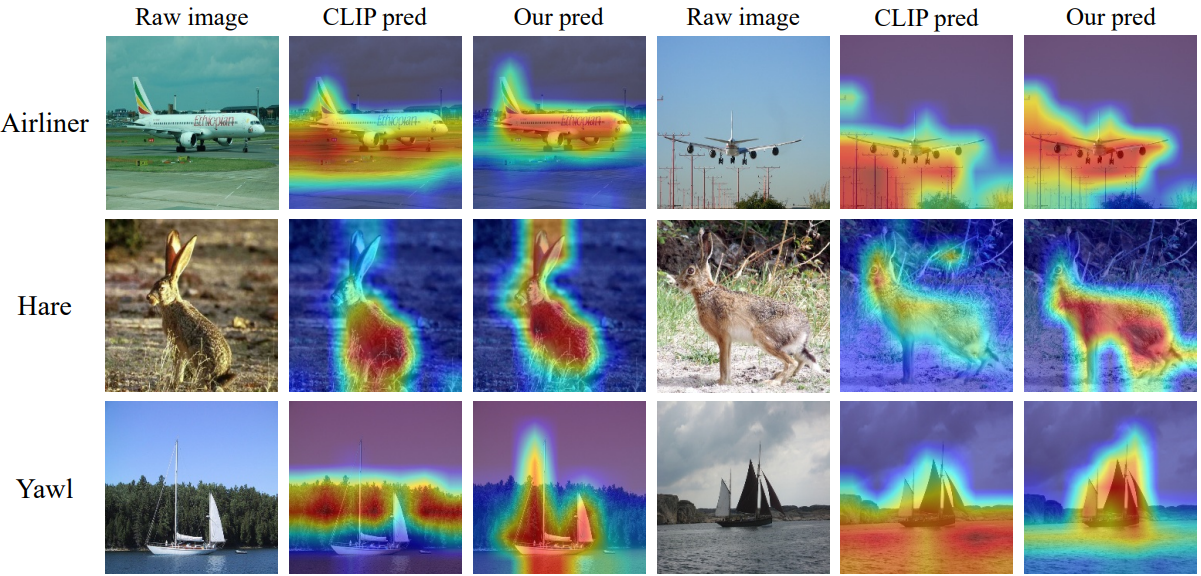}
\caption{Class activation maps for ALIP and CLIP on different classes from ImageNet.}
\label{fig6}
\vspace{-5mm}
\end{figure}

\noindent{\bf Effectiveness across Different Pre-training Datasets.}
In addition to YFCC15M, we conduct experiments on randomly selected subsets of 10M and 30M from LAION400M. For a more comprehensive comparison, we report the linear probe performance on 27 downstream datasets. As illustrated in Fig.~\ref{fig:laion}, ALIP significantly improves the performance on different models and pre-training datasets. Specifically, the ALIP-ViT-B/32 models pre-trained on LAION10M and LAION30M outperform the CLIP-ViT-B/32 models on 23 and 24 datasets, respectively. Additionally, when training a larger model, the ALIP-ViT-B/16 model surpasses the CLIP-ViT-B/16 model on 23 datasets. The experimental results demonstrate that ALIP exhibits robustness and extensibility. Please refer to the supplementary material for more detailed experimental results.

As shown in Fig.~\ref{fig6}, we compare the class activation maps of ALIP and CLIP on different classes from ImageNet. 
Here, we use the class label as the textual tokens.
As can be seen, ALIP is superior in aligning the image patches and textual tokens. For instance, CLIP only focuses on the body of the rabbit, but ALIP is able to also capture the  ears. These results highlight the potential of ALIP to enhance the performance of image-text retrieval tasks.

\section{Conclusion}
In this paper, we introduce a bi-path adaptive contrastive learning model, which includes the language consistency gate and description consistency gate. Specifically, LCG and DCG can adjust the weights of samples and image-text/caption pairs during training, thus effectively reducing the impact of the noisy or unaligned language description. Our method shows superior performance with different models and pre-training datasets on different downstream tasks. We hope our work could bring insights into exploiting the language-image pre-training model.

{\small
\bibliographystyle{iccv_2023}
\bibliography{iccv23_ALIP}
}

\clearpage

\appendix
\section{More Analysis}

\subsection{Ablation on Different Caption Model}
In Tab.~\ref{tab:1}, we present the performance using different caption models OFA vs. BLIP. The BLIP model we use is BLIP$_{CapFilt-L}$. Experimental results show that ALIP$_{BLIP}$ is on par with ALIP$_{OFA}$.

\begin{table}[h]
\caption{Comparison with different caption models.}
\label{tab:1}
\centering
\resizebox{0.9\columnwidth}{!}{
    \begin{sc}
    \begin{tabular}{lcccc}
    \toprule
     &\multicolumn{2}{c}{MSCOCO}  & \multicolumn{1}{c}{LINEAR PROBE} & \multicolumn{1}{c}{ZERO-SHOT} \\ 
     Methods & I2T    & T2I      &  \shortstack{Average} &  \shortstack{Average} \\ 
    \midrule
    ALIP$_{OFA}$  & 46.8  & 29.3  & 72.2  & 41.7  \\
    ALIP$_{BLIP}$  & 45.6  & 27.2  & 72.4  & 40.6  \\
    \bottomrule
    \end{tabular}
    \end{sc}
}
\vspace{-4mm}
\end{table}

\subsection{Analysis on Additional Costs}
ALIP-ViT-B/32 consumes 40\% extra FLOPs and 30\% more memory than CLIP-ViT-B/32. As the model size escalates, these extra requirements become less significant. ALIP-ViT-L/14 necessitates only about a 7\% increase in FLOPs and a mere 5\% additional memory, compared to CLIP-ViT-L/14.

\subsection{Analysis on Proportion of Noisy Pairs}
To better demonstrate the effectiveness of the ALIP, we conducted a statistical analysis on the number of data pairs where $W_{c}>W_{t}$. There are 4,153,279 pairs, which accounts for about 27.6\% of the total dataset.

\section{Detail Experimental Settings}

\subsection{Experimental Settings}
We show the settings in Tab.~\ref{tab:hyperparams} for ALIP pre-training.

\begin{table}[h]
    \centering
    \caption{Hyperparameters used for ALIP pre-training.}
    \label{tab:hyperparams}
    \centering
    \resizebox{0.6\linewidth}{!}{
     \begin{tabular}{l|c}
        \toprule Hyperparameter & Value \\
        \midrule
        Initial temperature & $0.07$ \\
        Adam $\beta_{1}$ & $0.9$ \\
        Adam $\beta_{2}$ & $0.98$ \\
        Adam $\epsilon$  & $10^{-6}$ \\
        Weight decay & $0.2$ \\
        Batch size & 4096 \\
        Learning rate & 0.001 \\
        Learning rate scheduler & OneCycleLR \\
        Pct start & 0.1 \\
        Training epochs & 32  \\
        GPU & $16 \times $V100 \\
        \bottomrule
    \end{tabular}
    }
\vspace{-4mm}
\end{table}

\subsection{Model Architectures}
We follow the same architecture as CLIP. Tab.~\ref{tab:alip_model_Hyperparameter} describe the detail of the ALIP-ViT-B/32 and ALIP-ViT-B/16.

\begin{table*}[h]
\caption{The architecture parameters for ALIP models.}
\label{tab:alip_model_Hyperparameter}
    \centering
    \resizebox{0.9\linewidth}{!}{
    \begin{sc}
    \begin{tabular}{l|cc cccc ccc}
        \toprule 
        \multirow{2}{*}{ Model } & Embedding & Input & \multicolumn{4}{c}{ Image Encoder } & \multicolumn{3}{c}{ Text Encoder } \\
          & dimension & resolution & layers & width & heads & patchs & layers & width & heads \\
        \midrule 
        ALIP-ViT-B/32  &  512 & $224\times 224$ & 12 & 768 & 12 & 32 & 12 & 512 & 8 \\
        ALIP-ViT-B/16 &  512 & $224\times 224$ & 12 & 768 & 12 & 16 & 12 & 512 & 8 \\
        \bottomrule
    \end{tabular}
    \end{sc}
    }
\end{table*}

\subsection{Prompts for Zero-shot Classification}
In this work, we evaluate the zero-shot performance of ALIP on 11 downstream datasets. All the prompts for the 11 downstream datasets are presented in Tab.~\ref{tab:prompt}.

\begin{table*}
\caption{Full list of prompts to evaluate the performance of zero-shot classification on 11 visual recognition datasets.}
\label{tab:prompt}
\centering
\resizebox{0.9\linewidth}{!}{
\begin{tabular}{llll}
\toprule
\multicolumn{4}{l}{\bf CIFAR 10 \& CIFAR 100} \\
a photo of a \{label\}. &
a blurry photo of a \{label\}. &
a black and white photo of a \{label\}. &
a low contrast photo of a \{label\}. \\
a high contrast photo of a \{label\}. &
a bad photo of a \{label\}. &
a good photo of a \{label\}. &
a photo of a small \{label\}. \\
a photo of a big \{label\}.&
a photo of the \{label\}.&
a blurry photo of the \{label\}.&
a black and white photo of the \{label\}. \\
a low contrast photo of the \{label\}.&
a high contrast photo of the \{label\}.&
a bad photo of the \{label\}.&
a good photo of the \{label\}. \\
a photo of the small \{label\}.&
a photo of the big \{label\}.& & \\
\midrule
\multicolumn{4}{l}{\bf Food101} \\
a photo of \{label\}, a type of food. & & \\
\midrule
\multicolumn{4}{l}{\bf Caltech101} \\
a photo of a \{label\}. &
a painting of a \{label\}. &
a plastic \{label\}. &
a sculpture of a \{label\}. \\
a sketch of a \{label\}. &
a tattoo of a \{label\}. &
a toy \{label\}. &
a rendition of a \{label\}. \\
a embroidered \{label\}. &
a cartoon \{label\}. &
a \{label\} in a video game. &
a plushie \{label\}. \\
a origami \{label\}. &
art of a \{label\}. &
graffiti of a \{label\}. &
a drawing of a \{label\}. \\
a doodle of a \{label\}. &
a photo of the \{label\}. &
a painting of the \{label\}.&
the plastic \{label\}. \\
a sculpture of the \{label\}.&
a sketch of the \{label\}.&
a tattoo of the \{label\}.&
the toy \{label\}. \\
a rendition of the \{label\}.&
the embroidered \{label\}.&
the cartoon \{label\}.&
the \{label\} in a video game. \\
the plushie \{label\}.&
the origami \{label\}.&
art of the \{label\}.&
graffiti of the \{label\}. \\
a drawing of the \{label\}.&
a doodle of the \{label\}.& & \\
\midrule
\multicolumn{4}{l}{\bf Stanford Cars} \\
a photo of a \{label\}.&
a photo of the \{label\}.&
a photo of my \{label\}.&
i love my \{label\}! \\
a photo of my dirty \{label\}.&
a photo of my clean \{label\}.&
a photo of my new \{label\}.&
a photo of my old \{label\}. \\
\midrule
\multicolumn{4}{l}{\bf DTD} \\
a photo of a \{label\} texture.&
a photo of a \{label\} pattern.&
a photo of a \{label\} thing.&
a photo of a \{label\} object. \\
a photo of the \{label\} texture. &
a photo of the \{label\} pattern. &
a photo of the \{label\} thing. &
a photo of the \{label\} object. \\
\midrule
\multicolumn{4}{l}{\bf FGVC Aircraft} \\
a photo of a \{label\}, a type of aircraft.&
a photo of the \{label\}, a type of aircraft.& & \\
\midrule
\multicolumn{4}{l}{\bf Flowers102} \\
a photo of a \{label\}, a type of flower. &&& \\
\midrule
\multicolumn{4}{l}{\bf Pets } \\
a photo of a \{label\}, a type of pet.&&& \\
\midrule
\multicolumn{4}{l}{\bf  SUN39} \\
a photo of a \{label\}.&
a photo of the \{label\}.&& \\
\midrule
\multicolumn{4}{l}{\bf  ImageNet} \\
a bad photo of a \{label\}. & 
a photo of many \{label\}. &
a sculpture of a \{label\}. &
a photo of the hard to see \{label\}. \\
a low resolution photo of the \{label\}. & 
a rendering of a \{label\}. &
graffiti of a \{label\}. &
a bad photo of the \{label\}.  \\
a cropped photo of the \{label\}. &
a tattoo of a \{label\}. & 
the embroidered \{label\}. &
a photo of a hard to see \{label\}.  \\
a bright photo of a \{label\}.&
a photo of a clean \{label\}.&
a photo of a dirty \{label\}.&
a dark photo of the \{label\}. \\
a drawing of a \{label\}.&
a photo of my \{label\}.&
the plastic \{label\}.&
a photo of the cool \{label\}. \\
a close-up photo of a \{label\}.&
a black and white photo of the \{label\}.&
a painting of the \{label\}.&
a painting of a \{label\}. \\
a pixelated photo of the \{label\}.& 
a sculpture of the \{label\}.&
a bright photo of the \{label\}.&
a cropped photo of a \{label\}. \\
a plastic \{label\}.&
a photo of the dirty \{label\}.& 
a jpeg corrupted photo of a \{label\}.&
a blurry photo of the \{label\}. \\
a photo of the \{label\}.&
a good photo of the \{label\}.&
a rendering of the \{label\}.&
a \{label\} in a video game. \\
a photo of one \{label\}.&
a doodle of a \{label\}.&
a close-up photo of the \{label\}.&
a photo of a \{label\}. \\
the origami \{label\}.&
the \{label\} in a video game.&
a sketch of a \{label\}.&
a doodle of the \{label\}. \\
a origami \{label\}.&
a low resolution photo of a \{label\}.&
the toy \{label\}.&
a rendition of the \{label\}. \\
a photo of the clean \{label\}.& 
a photo of a large \{label\}.& 
a rendition of a \{label\}.&
a photo of a nice \{label\}. \\
a photo of a weird \{label\}.& 
a blurry photo of a \{label\}.&
a cartoon \{label\}.&
art of a \{label\}. \\
a sketch of the \{label\}.& 
a embroidered \{label\}.&
a pixelated photo of a \{label\}.&
itap of the \{label\}. \\
a jpeg corrupted photo of the \{label\}.& 
a good photo of a \{label\}.&
a plushie \{label\}.&
a photo of the nice \{label\}. \\
a photo of the small \{label\}.& 
a photo of the weird \{label\}.&
the cartoon \{label\}.&
art of the \{label\}. \\
a drawing of the \{label\}.& 
a photo of the large \{label\}.& 
a black and white photo of a \{label\}.&
the plushie \{label\}. \\
a dark photo of a \{label\}.& 
itap of a \{label\}.& 
graffiti of the \{label\}.& 
a toy \{label\}. \\
itap of my \{label\}.& 
a photo of a cool \{label\}.&
a photo of a small \{label\}.& 
a tattoo of the \{label\}. \\
\bottomrule
\end{tabular}}
\end{table*}

\section{Detail Linear Probe on LAION}
\subsection{Downstream Datasets}
We use 26 image classification datasets to prove the effectiveness of our method. These datasets
include Food101~\cite{bossard2014food}, CIFAR10~\cite{krizhevsky2009learning}, CIFAR100~\cite{krizhevsky2009learning}, 
Birdsnap~\cite{berg2014birdsnap},
SUN397~\cite{xiao2010sun},
Stanford Cars~\cite{KrauseStarkDengFei-Fei_3DRR2013},
FGVC Aircraft~\cite{maji2013fine},
VOC2007~\cite{everingham2007pascal},
DTD~\cite{cimpoi2014describing},
Pets~\cite{parkhi2012cats}, 
Caltech101~\cite{fei2004learning},
Flowers102~\cite{nilsback2008automated},
MNIST~\cite{lecun1998gradient},
SLT10~\cite{coates2011analysis},
EuroSAT~\cite{helber2019eurosat},
RESISC45~\cite{cheng2017remote},
GTSRB~\cite{stallkamp2012man},
KITTI~\cite{geiger2012we},
Country211~\cite{radford2021learning},
PCAM~\cite{veeling2018rotation},
UCF101~\cite{soomro2012ucf101},
Kinetics700~\cite{carreira2019short},
CLEVR~\cite{johnson2017clevr},
Hateful Memes~\cite{kiela2020hateful},
SST2~\cite{radford2021learning},
ImageNet~\cite{deng2009imagenet}.Details on each dataset and the corresponding evaluation metrics are provided in Tab.~\ref{linearprobedatasets}.

\begin{table*}[h]
\caption{List of linear probe datasets with the data distribution and evaluation metrics.}
\label{linearprobedatasets}
\centering
\resizebox{0.65\linewidth}{!}{
\begin{tabular}{lcccr}
\toprule
\multicolumn{1}{l}{Dataset} & \multicolumn{1}{c}{Classes} & \multicolumn{1}{c}{Train size} & \multicolumn{1}{c}{Test size} & \multicolumn{1}{c}{Evaluation metric} \\
\midrule
Food101                        & 102                             & 75,750                             & 25,250                            & accuracy                                  \\
CIFAR10                        & 10                              & 50,000                             & 10,000                            & accuracy                                  \\
CIFAR100                       & 100                             & 50,000                             & 10,000                            & accuracy                                  \\
Birdsnap                        & 500                             & 42,138                             & 2,149                             & accuracy                                  \\
SUN397                          & 397                             & 19,850                             & 19,850                            & accuracy                                  \\
Cars                   & 196                             & 8,144                              & 8,041                             & accuracy                                  \\
Aircraft                   & 100                             & 6,667                              & 3,333                             & mean per class                            \\
VOC2007                   & 20                             & 5011                             & 4952                             & 11-point mAP                            \\
DTD           & 47                              & 3,760                              & 1,880                             & accuracy                                  \\
Pets                & 37                              & 3,680                              & 3,669                             & mean per class                            \\
Caltech101                     & 101                             & 3,000                              & 5,677                             & mean-per-class                            \\
Flowers                  & 102                             & 2,040                              & 6,149                             & mean per class                            \\
MNIST                  & 10                             & 60,000                              & 10,000                            & accuracy                            \\
STL10                        & 10                             & 5,000                             & 8,000                            & accuracy                                  \\
EuroSAT                         & 10                              & 10,000                             & 5,000                             & accuracy                                  \\
RESISC45            & 45                              & 3,150                              & 25,200                             & accuracy                                  \\
GTSRB            & 43                 &26,640                 &12,630                                                   & accuracy                                  \\
KITTI            & 4                              & 6770                             & 711                             & accuracy                                  \\
Country211            & 211                              & 42,200                             & 21,100                             & accuracy                                  \\
PCAM            & 2                              & 294,912                             & 32,768                             & accuracy                                  \\
UCF101            & 101                              & 9,537                             & 1,794                            & accuracy                                  \\
Kinetics700            & 700                              & 530,779                             & 33,944                            & mean(top1,top5)                                  \\
CLEVR            & 8                              & 2,000                             & 500                            & accuracy                                  \\
Memes            & 2                              & 8,500                             & 500                            & ROC AUC                                  \\
SST2            & 2                              & 7,792                             & 1,821                            & accuracy                                  \\
ImageNet                        & 1000                            & 1,281,167                          & 50,000                            & accuracy                                  \\
\bottomrule
\end{tabular}}
\end{table*}

\subsection{Detail Linear Probe results}
We conduct experiments on randomly selected subsets of 10M and 30M from the LAION400M dataset. To provide a comprehensive comparison, we report the performance of the linear probe on 26 downstream datasets, the complete experimental results are shown in Tab.~\ref{tab:linear-probe-big-table}. The experimental results indicate that ALIP demonstrates both robustness and extensibility.

\begin{table*}[h]
	\caption{Top-1 accuracy(\%) of linear probe on 26 image classification datasets. }
	\label{tab:linear-probe-big-table}
	\centering
	\resizebox{1.0\linewidth}{!}{
        \begin{sc}
		\begin{tabular}{llcccccccccccccccccccccccccccc}
			\toprule
			 & Method              & \shortstack{Pre-train\\ data} & \rotatebox[origin=lb]{90}{\smash{Food101}} & \rotatebox[origin=lb]{90}{\smash{CIFAR10}} & \rotatebox[origin=lb]{90}{\smash{CIFAR100}} & \rotatebox[origin=lb]{90}{\smash{Birdsnap}} & \rotatebox[origin=lb]{90}{\smash{SUN397}} & \rotatebox[origin=lb]{90}{\smash{Cars}} & \rotatebox[origin=lb]{90}{\smash{Aircraft}} & \rotatebox[origin=lb]{90}{\smash{VOC2007}} & \rotatebox[origin=lb]{90}{\smash{DTD}} & \rotatebox[origin=lb]{90}{\smash{Pets}} & \rotatebox[origin=lb]{90}{\smash{Caltech101}} & \rotatebox[origin=lb]{90}{\smash{Flowers}} & \rotatebox[origin=lb]{90}{\smash{MNIST}} & \rotatebox[origin=lb]{90}{\smash{STL10}} & \rotatebox[origin=lb]{90}{\smash{EuroSAT}} & \rotatebox[origin=lb]{90}{\smash{RESISC45}} & \rotatebox[origin=lb]{90}{\smash{GTSRB}} & \rotatebox[origin=lb]{90}{\smash{KITTI}} & \rotatebox[origin=lb]{90}{\smash{Country211}} & \rotatebox[origin=lb]{90}{\smash{PCAM}} & \rotatebox[origin=lb]{90}{\smash{UCF101}} & \rotatebox[origin=lb]{90}{\smash{Kinetics700}} & \rotatebox[origin=lb]{90}{\smash{CLEVR}} & \rotatebox[origin=lb]{90}{\smash{Memes}} & \rotatebox[origin=lb]{90}{\smash{SST2}} &
             \rotatebox[origin=lb]{90}{\smash{ImageNet}} &
             \rotatebox[origin=lb]{90}{\smash{Average}} \\

              \midrule
              & CLIP-ViT B/32     & LAION10M & 66.9	& 91.2	& 74.8	& 33.1	& 63.0 	& 71.1 	& 40.3 	& 80.9 	& 68.5 	& 71.0 	& 84.7 	& 89.5 	& 98.0 	& 93.6 	& 95.7 	& 78.4 	& 78.9 	& 72.0 	& 12.7 	& 83.2 	& 70.1 	& 41.5 	& 49.0 	& 53.8 	& 56.4 	& 54.8 & 68.2   \\    

              & ALIP-ViT B/32     & LAION10M & 71.5 	& 92.2 	& 76.1 	& 36.3 	& 67.3 	& 70.1 	& 41.8 	& 85.3 	& 71.3 	& 74.3 	& 86.9 	& 90.7 	& 98.0 	& 94.6 	& 95.4 	& 84.3 	& 84.1 	& 70.0 	& 12.9 	& 83.4 	& 75.9 	& 46.4 	& 51.0 	& 54.8 	& 56.5 	& 59.6 & 70.4   \\   \hline

               & CLIP-ViT B/16    & LAION10M & 74.2 	& 91.6 	& 76.2 	& 44.1 	& 65.5 	& 80.5 	& 42.9 	& 83.2 	& 70.0 	& 74.5 	& 85.5 	& 92.8 	& 98.2 	& 94.5 	& 96.2 	& 85.0 	& 79.2 	& 70.5 	& 14.9 	& 85.4 	& 75.5 	& 44.9 	& 49.0 	& 55.0 	& 58.3 	& 60.8  & 71.1 \\    

              & ALIP-ViT B/16     & LAION10M & 77.2 	& 93.3 	& 77.0 	& 45.1 	& 69.4 	& 77.3 	& 48.6 	& 87.7 	& 74.5 	& 79.0 	& 88.1 	& 93.0 	& 98.3 	& 96.3 	& 96.3 	& 86.4 	& 83.7 	& 72.2 	& 14.2 	& 85.2 	& 80.1 	& 50.1 	& 55.4 	& 55.7 	& 57.3 	& 64.8  & 73.3  \\         \hline          
  

              & CLIP-ViT B/32     & LAION30M & 73.1 	& 94.1 	& 79.6 	& 40.9 	& 66.4 	& 79.4 	& 41.5 	& 83.3 	& 71.6 	& 76.7 	& 87.4 	& 92.4 	& 97.8 	& 95.2 	& 95.3 	& 82.6 	& 82.3 	& 72.2 	& 14.6 	& 82.7 	& 73.0 	& 45.7 	& 44.0 	& 54.3 	& 57.8 	& 59.8  & 70.9  \\    

              & ALIP-ViT B/32     & LAION30M & 76.7 	& 94.0 	& 79.3 	& 44.2 	& 70.6 	& 77.7 	& 48.4 	& 87.6 	& 74.4 	& 80.4 	& 90.0 	& 93.8 	& 98.3 	& 96.3 	& 96.0 	& 86.7 	& 84.7 	& 72.3 	& 15.0 	& 85.0 	& 81.0 	& 50.6 	& 55.6 	& 56.1 	& 59.8 	& 65.0  & 73.8  \\  
			\bottomrule
		\end{tabular}
  \end{sc}}
\end{table*}

\section{More Visualization}
\subsection{Sample Visualization}
\begin{figure*}[t]
\centering
\includegraphics[width=0.85\linewidth]{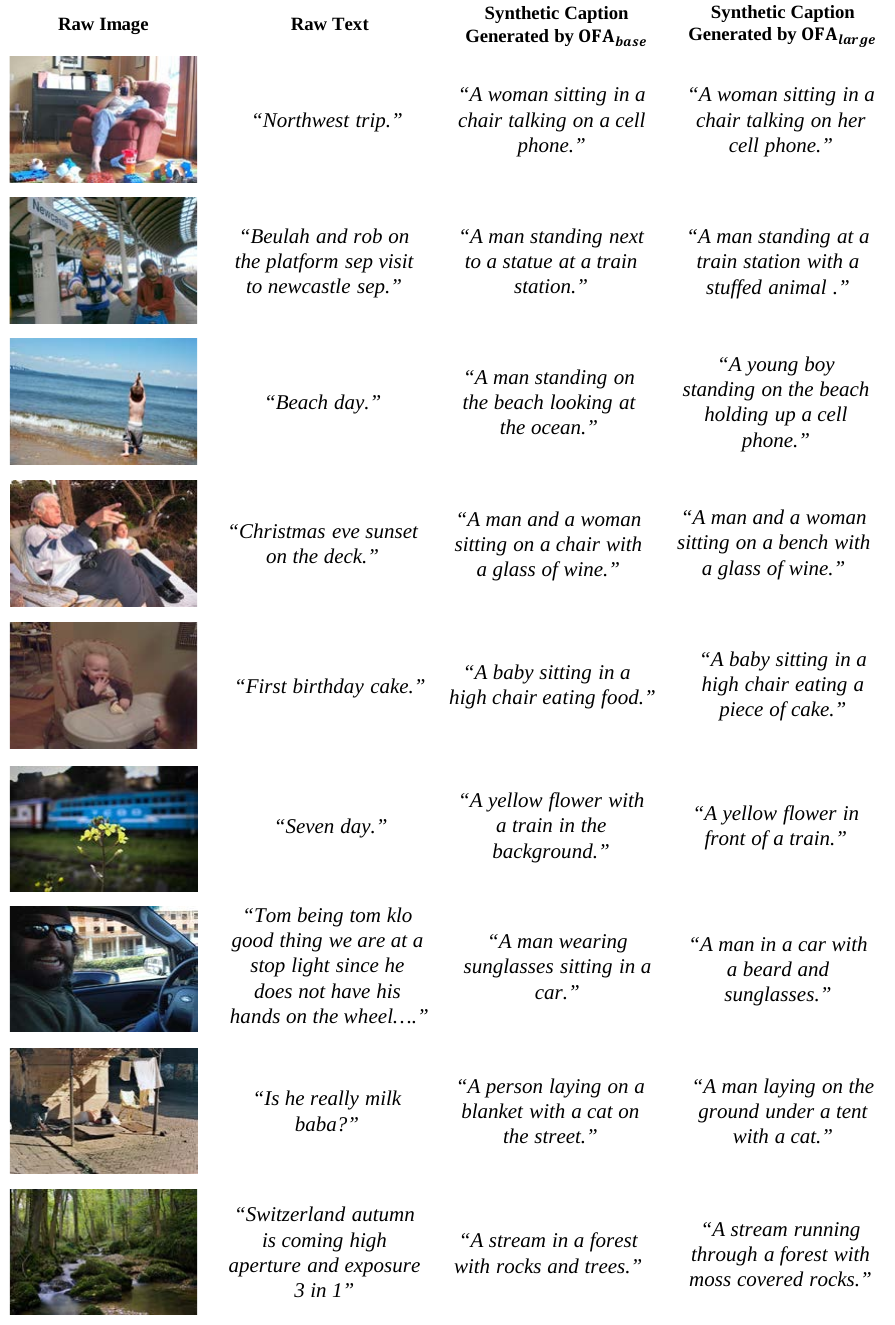}
\caption{Examples of the image-text-caption triplet pairs from YFCC15M. We present the synthetic captions generated by the OFA$_{base}$ and OFA$_{large}$.}
\vspace{-4mm}
\label{fig:casestudy}
\end{figure*}

In Fig.~\ref{fig:casestudy}, we present visualizations of samples with raw images, raw texts, synthetic captions generated by OFA$_{base}$, and synthetic captions generated by OFA$_{large}$. It can be observed that the synthetic captions contain supplementary information that can potentially enhance representation learning. Moreover, the captions generated by OFA$_{base}$ and OFA$_{large}$ exhibit minimal differences.

\subsection{Class Activation Maps}
\begin{figure*}[t]
\centering
\includegraphics[width=1\linewidth]{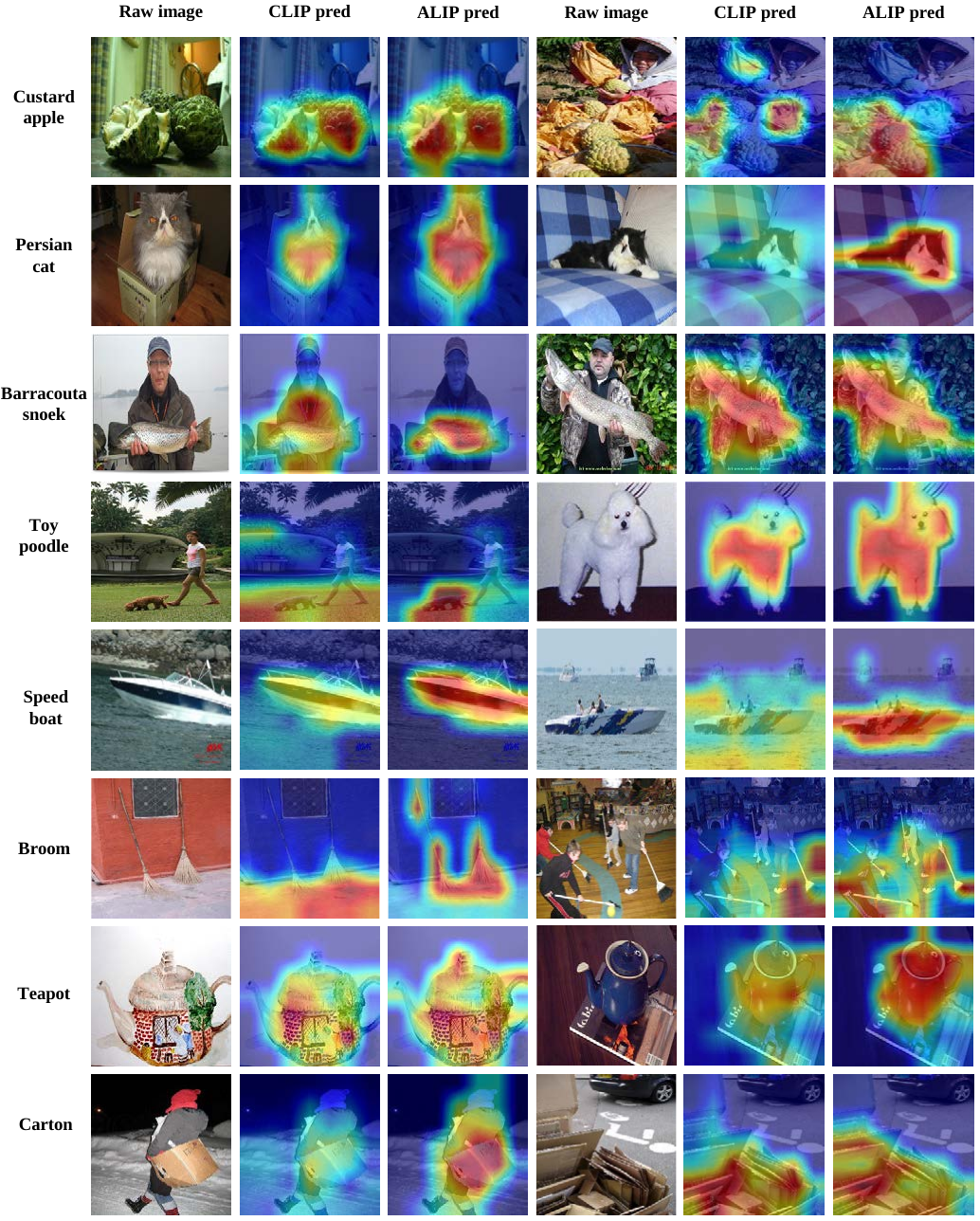}
\caption{More class activation maps for CLIP and ALIP on different classes from ImageNet.}
\vspace{-4mm}
\label{fig:classactivate}
\end{figure*}

In Fig.~\ref{fig:classactivate}, we present additional class activation maps of ALIP and CLIP for different classes from ImageNet. The visualizations demonstrate that ALIP is superior in effectively aligning image patches and textual tokens.

\end{document}